\documentclass[10pt,twocolumn,letterpaper]{article}

\usepackage[pagenumbers]{cvpr} 

\usepackage[dvipsnames]{xcolor}

\definecolor{cvprblue}{rgb}{0.21,0.49,0.74}
\usepackage[pagebackref,breaklinks,colorlinks,citecolor=cvprblue]{hyperref}
\usepackage{adjustbox}
\usepackage{color, colortbl}
\definecolor{beaublue}{rgb}{0.74, 0.83, 0.9}
\usepackage{hyperref}
\usepackage{multirow}
\usepackage[accsupp]{axessibility} 
\hypersetup{
    colorlinks=true,
    linkcolor=magenta,
    filecolor=magenta,      
    urlcolor=magenta,
}

\def\paperID{10585} 
\def\confName{CVPR}
\def\confYear{2024}

\title{
Accurate Spatial Gene Expression Prediction by Integrating Multi-Resolution Features
}

\author{Youngmin Chung, Ji Hun Ha, Kyeong Chan Im, Joo Sang Lee\thanks{Corresponding author.}\\
Sungkyunkwan University, South Korea\\
{\tt\small ymblue@g.skku.edu, joosang.lee@skku.edu}
}

\begin{document}
\maketitle
\begin{abstract}
Recent advancements in Spatial Transcriptomics (ST) technology have facilitated detailed gene expression analysis within tissue contexts. However, the high costs and methodological limitations of ST necessitate a more robust predictive model. In response, this paper introduces TRIPLEX, a novel deep learning framework designed to predict spatial gene expression from Whole Slide Images (WSIs). TRIPLEX uniquely harnesses multi-resolution features, capturing cellular morphology at individual spots, the local context around these spots, and the global tissue organization. By integrating these features through an effective fusion strategy, TRIPLEX achieves accurate gene expression prediction. Our comprehensive benchmark study, conducted on three public ST datasets and supplemented with Visium data from 10X Genomics, demonstrates that TRIPLEX outperforms current state-of-the-art models in Mean Squared Error (MSE), Mean Absolute Error (MAE), and Pearson Correlation Coefficient (PCC). The model's predictions align closely with ground truth gene expression profiles and tumor annotations, underscoring TRIPLEX's potential in advancing cancer diagnosis and treatment. 
\end{abstract}
    
\section{Introduction}
\label{sec:intro}

\hspace{\parindent} The emergence of large-scale Spatial Transcriptomics (ST) technology has facilitated the quantification of mRNA expression across a multitude of genes within the spatial context of tissue samples \cite{staahl2016visualization}. ST technology segments centimeter-scale Whole Slide Images (WSIs) into hundreds of thousands of small spots, each providing its gene expression profile. 
Considering the substantial cost associated with ST sequencing technology, coupled with the widespread availability of WSIs, a pressing question is how to best predict spatial gene expression based on WSIs using rapidly evolving computer vision techniques.


A number of studies have endeavored to address this challenge \cite{he2020integrating,pang2021leveraging,zeng2022spatial,yang2023exemplar,xie2023spatially}.
Approaches vary, with some predicting gene expression strictly from the tissue image confined within the spot's boundaries \cite{he2020integrating}, while others also take into account spatial dependencies between spot images \cite{pang2021leveraging,zeng2022spatial}, or consider similarities to reference spots \cite{yang2023exemplar,xie2023spatially}. However, we have noted several limitations inherent to these existing methodologies.
 Firstly, current methods primarily focus on spot images, neglecting the wealth of biological information available in the wider image context. By integrating both the specific spot and its surrounding environment, along with the holistic view of the entire histology image, we can access richer information, encompassing varied biological contexts.
Secondly, models that consider interactions between spots \cite{pang2021leveraging,zeng2022spatial} face a limitation in processing the embedding of all patches in a WSI simultaneously. This approach, common in handling hundreds to thousands of patches within a WSI, limits the scalability of the patch embedding model due to resource constraints. Such limitations significantly impede the extraction of fine-grained, rich representations from each spot, thereby affecting the model’s ability to perform detailed analysis of WSIs.
Thirdly, model performance is frequently overestimated because of inadequate validation, such as using the limited size of dataset \cite{he2020integrating} sometimes without cross-validation \cite{xie2023spatially} and training/testing with replicates from the same patient \cite{pang2021leveraging,zeng2022spatial,yang2023exemplar}.
The limited size of ST datasets means that exclusive reliance on a single dataset for model evaluation can hinder an accurate assessment of the model's capabilities, thereby emphasizing the necessity for cross-validation.
The issue is compounded when replicate data from the same patient, often featuring nearly identical image-gene expression pairs, are used in both training and testing phases. This can lead to an inflated perception of a model's effectiveness, as it may not accurately reflect the model's ability to generalize to new, unseen data.
Lastly, the use of disparate datasets, diverse normalization methods, and varied evaluation techniques in existing research studies compounds the challenge of conducting fair comparisons of the models.

Addressing these limitations, we present TRIPLEX, an innovative deep learning framework designed to leverage multi-resolution features from histology images for robustly predicting spatial gene expression levels. TRIPLEX extracts three distinct types of features corresponding to different resolutions: the target spot image (representing the specific spot, whose gene expression to be predicted), the neighbor view (encompassing a wider area around the spot), and the global view (comprising the aggregate of all spot images). These features capture varying levels of biological information—ranging from the detailed cell morphology in the target spot image, to the surrounding tissue phenotype, and the overall tissue microenvironment in the WSI. Each is integral to understanding the spatial gene expression levels of the given spot.
TRIPLEX employs separate encoders to extract these features from WSIs, each focusing on its assigned resolution to efficiently capture relevant details. For neighbour or global view with larger resolution, pre-extracted features are used to reduce the burden of computational cost, while for target spot images, encoders are fully updated to extract fine-grained information. These features are then integrated via a fusion layer for effective gene expression prediction. This approach allows TRIPLEX to utilize resolution-specific information, thereby enhancing prediction accuracy while avoiding significant increases in computational costs.

Our study sets a new benchmark in spatial gene expression prediction, comparing our model, TRIPLEX, against five prior studies \cite{he2020integrating,pang2021leveraging,zeng2022spatial,yang2023exemplar,xie2023spatially} under uniform experimental conditions. We conduct internal evaluations using three public Spatial Transcriptomics (ST) datasets \cite{andersson2020spatial,ji2020multimodal,he2020integrating} and external validations using higher-resolution Visium data from 10X Genomics. Our validation procedure strictly avoids mixing patient sample replicates between training and testing datasets, a significant departure from previous methods \cite{pang2021leveraging,zeng2022spatial,yang2023exemplar}, and employs rigorous cross-validation. Our results indicate that TRIPLEX surpasses existing models in terms of Mean Squared Error (MSE), Mean Absolute Error (MAE), and Pearson Correlation Coefficient (PCC) in both internal and external evaluations. Furthermore, we provide visualizations of the expression distributions for a specific gene commonly associated with cancer. These visualizations reveal that our model’s predictions align more closely with actual gene expression data and tumor annotations, demonstrating its enhanced predictive accuracy.

Our key contributions can be summarized as follows:

\begin{itemize}
\item We introduce an innovative approach to predict spatial gene expression levels from WSIs by integrating multiple biological contexts. 

\item Our proposed framework seamlessly integrates multi-resolution features. This integration is facilitated by a feature extraction strategy, the use of various types of transformers, and a fusion loss technique, all while keeping the additional computational costs to a minimum.

\item Through comprehensive experiments on three public ST datasets and additional external evaluations using three Visium data, our study establishes a new benchmark in the field of spatial gene expression prediction. The results consistently show that our proposed method outperforms all existing models included in our comparative analysis.
\end{itemize}

\section{Related Work}
\label{sec:Related Work}

\hspace{\parindent} In this section, we delve into studies pertinent to our research. For clarity, the term 'spot' will be used to denote a predefined unit region within a WSI where gene expression is quantified. Moreover, we will use 'target spot' to specifically refer to the spot within a WSI for which we seek to predict gene expression.

\noindent \textbf{Spatial gene expression prediction from WSIs via deep learning} 
We review the pioneering works in this field, which aim to predict spatial gene expression from WSIs. ST-Net \cite{he2020integrating} utilizes a standard transfer learning strategy, training a Densenet121 model \cite{huang2017densely}—pretrained on ImageNet—using histology images as input and gene expression as labels.
Following this, HisToGene \cite{pang2021leveraging} leverages Vision Transformers (ViT) \cite{dosovitskiy2020image} to account for correlations among patches in a WSI, thereby predicting gene expression from global-context aware features. Further developing this concept, Hist2ST \cite{zeng2022spatial} enhances the approach by emphasizing patch embedding using ConvMixer \cite{trockman2022patches} and aggregating neighborhood information through graph convolution network \cite{kipf2016semi}.
While these previous works \cite{pang2021leveraging,zeng2022spatial} share similarities with our methodology, they predominantly process patch and global embedding sequentially, often overlooking the neighboring information around the target spot. In contrast, our method sets itself apart by concurrently extracting critical features at three distinct resolutions, including the neighbor view, and integrating them for gene expression prediction.
In a different vein, EGN \cite{yang2023exemplar} adopts exemplar learning for predicting gene expression from histology images. This method dynamically selects the most analogous exemplars from a target spot within a WSI to enhance prediction accuracy. Additionally, BLEEP \cite{xie2023spatially} introduces a bi-modal embedding framework similar to CLIP \cite{radford2021learning} to co-embed spot images and gene expression. After training, this model imputes the gene expression of a query spot using the retrieved gene expression set from a reference dataset.

\noindent \textbf{Deep learning for WSIs} 
Due to their gigapixel resolution, WSIs present a significant challenge for conventional deep learning frameworks in computer vision. To tackle this, Multiple Instance Learning (MIL) has been employed, enabling the handling of high-resolution data with sparse local annotations. In the context of WSIs, MIL approaches typically predict bag labels, such as distinguishing slides from cancer patients versus healthy individuals, by aggregating information from numerous small patches within the WSIs \cite{campanella2019clinical,kanavati2020weakly,li2021dual,shao2021transmil}. Recently, attention-based networks have been employed in MIL to aggregate all patches in WSIs, achieving state-of-the-art performance \cite{lu2021data,shao2021transmil,li2021dual}.
Moreover, Chen et al. \cite{chen2022scaling} introduced a Hierarchical Image Pyramid Transformer (HIPT) to adapt Vision Transformers (ViT) for WSIs. They effectively captured the hierarchical structures of WSIs by sequentially training ViTs across images of various resolutions, employing a self-supervised learning approach. This method has shown superior results in cancer subtyping and survival prediction, surpassing previous models. Drawing inspiration from this, our proposed method is specifically designed to simultaneously handle information from these multi-resolution features for predicting spatial gene expression.

\section{Method}
\label{sec:Method}

\begin{figure*}[!t]
    \centering
    \includegraphics[width=1\linewidth]{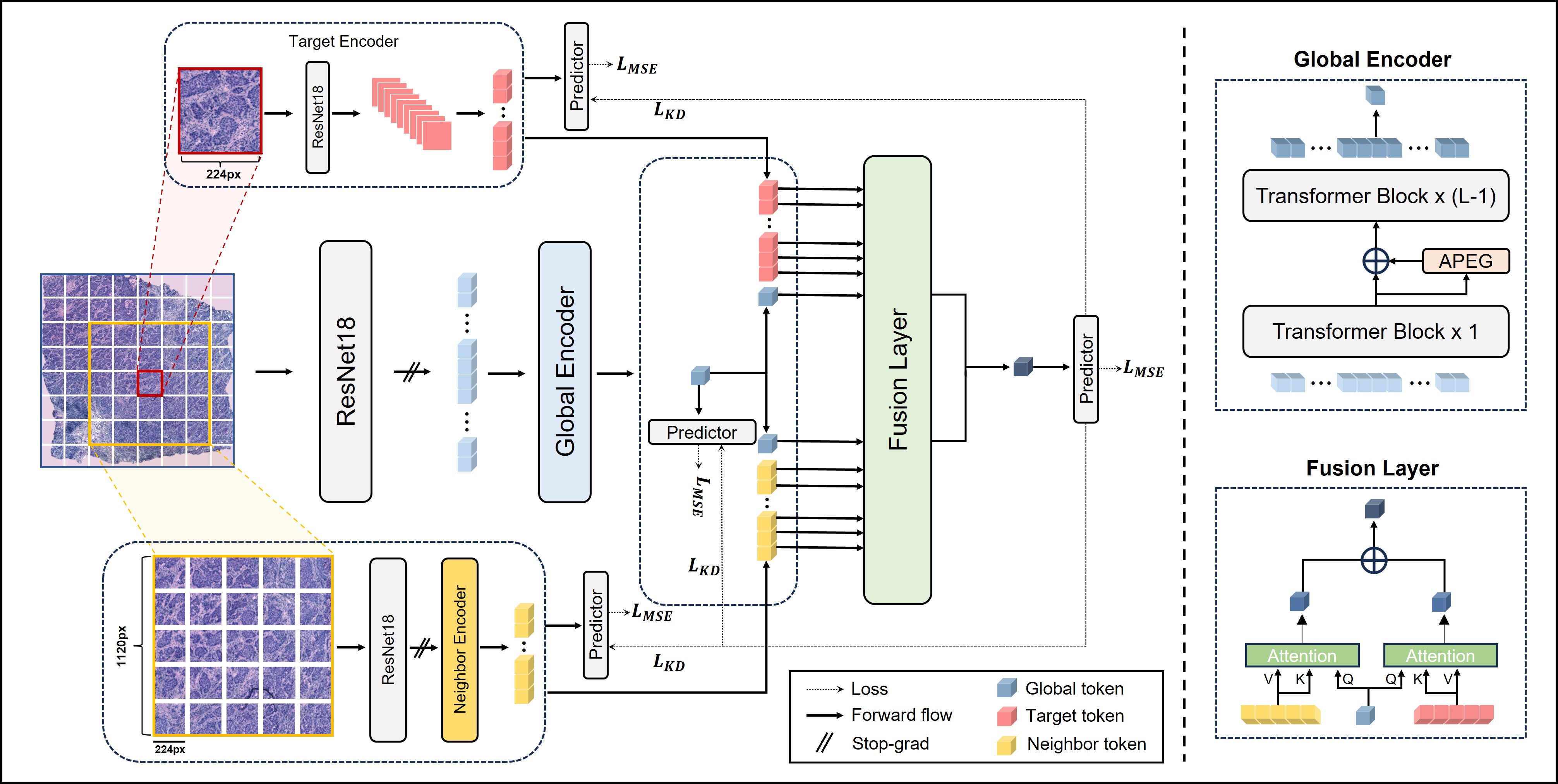}
    \caption{Schematic representation of the TRIPLEX. The global encoder processes the global view, while separate encoders handle the target spot image and neighbor view. A fusion layer, incorporated with fusion loss, facilitates the effective integration of these tokens to predict gene expression levels.
    }
    \label{fig:Fig2}
\end{figure*}

\subsection{Preliminary}
\hspace{\parindent} In this section, we present our problem formulation and detail our proposed method, concurrently assessing it in contrast to previous methodologies. Our task is a multi-output regression problem, where we input a set of spot images from a WSI, denoted as $\mathbf{X} \in \mathbb{R}^{n \times H \times W \times 3}$, and aim to predict the gene expression levels of individual spots, represented as $\mathbf{Y} \in \mathbb{R}^{n \times m}$. Here, $n$ denotes the number of spot images in a WSI, $m$ represents the number of genes whose expression levels to be predicted, and $H$ and $W$ signify the height and width of each spot image, respectively.

ST-Net \cite{he2020integrating} approaches the prediction task by estimating $\hat{Y}_i\in \mathbb{R}^{m}$ based solely on the information from a single spot image $X_i \in \mathbb{R}^{H \times W \times 3}$. This is represented as:
\begin{equation}
\hat{Y}_{i} = f(X_{i}) \in \mathbb{R}^{m}
\end{equation} 
where $\textit{i}$ indexes a target spot within a WSI. ST-Net is formulated under general supervised learning principles, where each input corresponds to a unique label, and each input is treated independently.

On the other side, methods employing ViT \cite{pang2021leveraging,zeng2022spatial} predict the gene expression of all spot images concurrently, expressed as:
\begin{equation}
\hat{Y} = f(X_{1}, X_{2}, ... ,X_{n}) \in \mathbb{R}^{n \times m}
\end{equation}
In this case, the problem is defined within the context of supervised learning, but with a key difference: the inputs are interdependent, affecting the prediction outcome collectively.

EGN \cite{yang2023exemplar} approaches the problem from a different perspective and predicts $\hat{Y}_i$ using $X_i$ together with its global view $G_i$ and its exemplar set $\{g_j, y_j\}_{j=1}^k$ $\in$ $K_i$. This can be formulated as:
\begin{equation}
\hat{Y}_{i} = f(X_{i}, G_i, K_i) \in \mathbb{R}^{m}
\end{equation}
where $G_i$ represents the features extracted from the target spot image $X_i$ using a pretrained model. The set $K_i$ comprises the k-nearest global views to $G_i$ and their associated gene expression levels.

In BLEEP \cite{xie2023spatially}, a bi-modal pretraining phase is leveraged, employing contrastive loss \cite{radford2021learning} to generate embeddings for both images and gene expression levels. During the inference phase, a given query image $X_i$ is processed through an image encoder, $Enc^{img}$, to produce its d-dimensional embedding vector $v_i$:
\begin{equation}
v_i = Enc^{img}(X_i) \in \mathbb{R}^{d}
\end{equation}
Simultaneously, an expression encoder, $Enc^{exp}$, is applied to the gene expression levels of the reference dataset, yielding an d-dimensional embedding vectors $e^{ref}$ represented as:
\begin{equation}
e^{ref} = Enc^{exp}(Y^{ref}) \in \mathbb{R}^{l \times d}
\end{equation}
Here, $l$ signifies the number of gene expression levels in the reference dataset. The process continues by identifying the top-k closest embeddings to $v_i$, denoted as $e^{top} \in \mathbb{R}^{k \times d}$. The corresponding gene expression values for these top-k embeddings, $Y^{top} \in \mathbb{R}^{k \times m}$, are then retrieved. The gene expression level for the query spot is estimated by computing the average of $Y^{top}$:
\begin{equation}
\hat{Y}_{i} = Average(Y^{top}) \in \mathbb{R}^{m}
\end{equation}

Our proposed method diverges from the previously mentioned approaches by employing a unique combination of three input types to predict $\hat{Y}_{i}$. These are the target spot image ${X^{Ta}_{i}}$, the local neighbor views ${X^{Ne}_{i}}$, and a collection of all the global views $G$: ${g_1,g_2,...,g{n}}$.

\begin{gather}
z^{Ta}_{i} = Enc^{Ta}(X^{Ta}_{i}) \in \mathbb{R}^{n^{Ta} \times d} \\
z^{Ne}_{i} = Enc^{Ne}({X^{Ne}_{i}}) \in \mathbb{R}^{n^{Ne} \times d} \\
z^{Gl} = Enc^{Gl}(G) \in \mathbb{R}^{n \times d} \\
\hat{Y}{i} = f(z^{Ta}_{i}, z^{Ne}_{i}, z^{Gl}_{i}) \in \mathbb{R}^{m}
\end{gather}
Here, $Enc^{Ta}$, $Enc^{Ne}$, and $Enc^{Gl}$ represent models that independently embed each type of input. The dimension of each embedded token is denoted by \textit{d}. The local neighbor views ${X^{Ne}_{i}}$ consist of the $n^{Ne}$ adjacent patches around the target spot image $X^{Ta}_i$, and $n^{Ta}$ signifies the number of tokens derived from $X^{Ta}_i$. 
In all cases, \textit{f} refers to a neural network function outputting gene expression levels.

\subsection{TRIPLEX}
\hspace{\parindent} The overall workflow of our method is illustrated in Figure \ref{fig:Fig2}. Initially, we process the global view, derived from all spot images of a WSI, through a global encoder to produce global tokens. Although these global tokens capture the macroscopic spatial distribution and inter-spot correlations, they might not adequately represent the detailed information specific to each target spot. To address this, we independently encode the target spot image and its neighboring views, generating tokens that encapsulate finer details. These are then integrated with the global tokens via fusion layers, enriching the global representation with specific target-related information.
Additionally, considering the diverse contextual information provided by different input sources, we have implemented a fusion loss mechanism, inspired by knowledge distillation, to enhance the efficacy of the fusion process.
Detailed descriptions of the individual components involved in our proposed model will be provided in the following sections.

\subsection{Embedding Global Information}
\hspace{\parindent} Processing all 224x224-sized images of spots starting from patch embedding is computationally intensive since the number of spots typically ranges from hundreds to thousands. To address this, we employ a feature extraction strategy commonly used in the MIL approach for WSIs. Specifically, we utilize a ResNet18 model pretrained on large-scale histology images for feature extraction \cite{ciga2022self}. The features thus obtained serve as input to the global encoder, which generates global tokens. This encoder comprises a series of transformer blocks and a Position Encoding Generator (PEG) \cite{chu2021conditional}, adept at encoding positional information for a variable number of spots.
Given that PEG was originally designed for natural images with regular, square shapes, we modify it for our specific use-case where the image shape is irregular. Our adaptation, termed the Atypical Position Encoding Generator (APEG), imbues tokens with absolute positional information pertinent to all spots in a WSI. This modification is crucial for effectively capturing the spatial distribution within the WSI.

\noindent \textbf{Atypical Position Encoding Generator (APEG)}
\hspace{\parindent} In our APEG framework, after the initial transformation of tokens through one transformer block, we employ a technique to re-establish their relative positional context. This involves reshaping the global tokens $z^{Gl} \in \mathbb{R}^{n \times d}$ into a spatial format $\hat{z}^{Gl} \in \mathbb{R}^{h \times w \times d}$. Here, $h$ and $w$ represent the maximum values of the $x$- and $y$-coordinates, respectively. During this process, any voids in the spatial arrangement are temporarily filled with zeros.
Subsequently, we apply convolutional layers to these reshaped tokens. This step includes refilling the areas that were previously vacant with zeros. After this convolutional processing, we revert the tokens back to their original format. This method enables us to effectively incorporate the relative spatial information of the tokens, enhancing the positional encoding within the WSI framework.

\subsection{Embedding Target/Neighbor Information}

\hspace{\parindent} We independently encode the image of the target spot and its neighboring view, generating tokens for both target and neighbor that are rich in contextual information. \\
\textbf{Embedding target information}
To encode the target spot image, we utilize a ResNet18 architecture, excluding the global average pooling and the fully-connected layer. This encoder processes each 224x224 target spot image by embedding it into 49 distinct features, with each feature having a dimension of 512. Notably, while this instance of ResNet18 is initialized with the same weights as before, it undergoes unique updates during the training process, ensuring tailored feature extraction for each target spot image. \\
\textbf{Embedding neighbor information}
For the neighbor view, we use the surrounding 1120x1120 image of each target spot. This choice is based on images directly adjacent to the target spot, rather than a group of neighboring spot images. This approach addresses the issue of non-uniform alignment and spacing between spots in ST data (refer to supplementary Figure 1 for details). In cases where a target spot is at the edge of the slide, zero padding is applied to maintain the required image size. This 1120x1120 neighbor view is then embedded into twenty-five 512-dimensional feature vectors using the ResNet18. These vectors then serve as input for the neighbor encoder.
The embedding process within the neighbor encoder involves a sequence of self-attention blocks, each integrated with relative position encoding \cite{shaw2018self}. Although the ResNet18 weights used for feature extraction remain fixed, the weights of the neighbor encoder are dynamically updated during training. Further details about the neighbor encoder are provided in the supplementary material.
\begin{table*}[ht]
\centering
\begin{adjustbox}{width=1\textwidth}
\label{t4}
\footnotesize
\begin{tabular}{c|c|ccc|ccc|ccc}
\noalign{\smallskip}\noalign{\smallskip}
\hline
\hline
     &  & \multicolumn{3}{c|}{BC1} & \multicolumn{3}{c|}{BC2} & \multicolumn{3}{c}{SCC} \\
   Source & Model & MSE & PCC(M) & PCC(H) & MSE & PCC(M) & PCC(H) & MSE & PCC(M) & PCC(H) \\
\hline
 & ST-Net \cite{he2020integrating} & $0.260\pm0.04$ & $0.194\pm0.11$ & $0.345\pm0.16$ &
$0.209\pm0.02$ & $0.116\pm0.06$ & $0.223\pm0.10$ &
$0.294\pm0.07$ & $0.274\pm0.08$ & $0.382\pm0.08$ \\
 & EGN \cite{yang2023exemplar} & $0.241\pm0.06$ & $0.197\pm0.11$ & $0.328\pm0.17$ &
$0.192\pm0.02$ & $0.111\pm0.05$ & $0.203\pm0.09$ &
$0.281\pm0.08$ & $0.281\pm0.06$ & $0.388\pm0.06$ \\
 \textbf{Local} & BLEEP \cite{xie2023spatially} & $0.277\pm0.05$ & $0.151\pm0.11$ & $0.277\pm0.16$ &
$0.235\pm0.02$ & $0.095\pm0.05$ & $0.193\pm0.10$ &
$0.297\pm0.05$ & $0.269\pm0.07$ & $0.396\pm0.08$ \\
 & TEM & $0.252\pm0.04$ & $0.208\pm0.11$ & $0.365\pm0.15$ &
$0.190\pm0.02$ & $0.119\pm0.06$ & $0.227\pm0.10$ &
$0.290\pm0.06$ & $0.296\pm0.07$ & $0.402\pm0.08$ \\
 & NEM & $0.278\pm0.08$ & $0.255\pm0.13$ & $0.424\pm0.18$ &
$0.193\pm0.03$ & $0.152\pm0.05$ & $0.277\pm0.09$ &
$0.373\pm0.14$ & $0.308\pm0.05$ & $0.444\pm0.06$ \\
\hline
 \textbf{Global} & HistoGene \cite{pang2021leveraging} & $0.314\pm0.09$ & $0.168\pm0.12$ & $0.302\pm0.19$ &
$0.194\pm0.05$ & $0.100\pm0.05$ & $0.219\pm0.12$ &
$0.270\pm0.09$ & $0.133\pm0.06$ & $0.261\pm0.13$ \\
 & GEM & $0.253\pm0.06$ & $0.295\pm0.14$ & $0.491\pm0.17$ &
$0.221\pm0.03$ & $0.193\pm0.06$ & $0.341\pm0.08$ &
$0.317\pm0.16$ & $0.276\pm0.09$ & $0.392\pm0.08$ \\
\hline
\textbf{Local+Global} & Hist2ST \cite{zeng2022spatial} & $0.285\pm0.08$ & $0.118\pm0.10$ & $0.248\pm0.17$ &
$\textbf{0.181}\pm0.02$ & $0.044\pm0.02$ & $0.099\pm0.03$ &
$1.291\pm0.65$ & $0.004\pm0.01$ & $0.053\pm0.01$ \\
\hline
\rowcolor{beaublue}
\textbf{Multiple} & TRIPLEX & $\textbf{0.228}\pm0.07$ & $\textbf{0.314}\pm0.14$ & $\textbf{0.497}\pm0.17$ &
$0.202\pm0.02$ & $\textbf{0.206}\pm0.07$ & $\textbf{0.352}\pm0.10$ &
$\textbf{0.268}\pm0.09$ & $\textbf{0.374}\pm0.07$ & $\textbf{0.490}\pm0.07$ \\
\hline
\hline
\end{tabular}
\end{adjustbox}
\caption{Cross validation result on each ST dataset. PCC(M) and PCC(H) denote the mean PCC for all genes and the mean PCC for highly predictive genes, respectively. The mean and standard deviation of cross-validation results are displayed. 
MAEs are excluded due to space limitations and can be found in the supplementary.
}
\label{table:table1}
\end{table*}
\subsection{Integrating global, neighbor, and target information}
To achieve an effective exchange of information and enhanced contextual understanding between different token types, we implement cross-attention layers. In this setup, the global token acts as the query (Q), with the target and neighboring tokens within the WSI forming the key (K) and value (V) pairs. These Q, K, and V components are utilized in a dot-product attention mechanism, which is crucial for assessing the relative importance of each target and neighbor token in comparison to the global token. This process is instrumental in developing a comprehensive understanding of the local and global contexts at each spot, thereby augmenting the model's performance in complex gene expression prediction tasks.
The integration of these insights is realized by summing the resultant tokens from the cross-attention layers:
\begin{gather}
z_i^{GT} = CrossAttn(z^{Gl}_i, z^{Ta}_i) \in \mathbb{R}^{d} \\
z_i^{GN} = CrossAttn(z^{Gl}_i, z^{Ne}_i) \in \mathbb{R}^{d} \\
z_i^{GTN} = Sum(z_i^{GT}, z_i^{GN}) \in \mathbb{R}^{d}
\end{gather}
In these equations, $z_i^{GT}$ and $z_i^{GN}$ represent the cross-attention results for the target and neighbor tokens, respectively, relative to the global token. $z_i^{GTN}$ is the summation of these two tokens, which is used to estimate the gene expression levels.
During the process, information exchange between the neighbor token and the target token occurs only through the global tokens. In this manner, we can efficiently integrate the three pieces of information at a minimal additional cost. \\
\textbf{Fusion Loss and Objective Function}
\hspace{\parindent} To optimize the integration of information from multiple tokens, we have introduced a fusion loss mechanism. This approach capitalizes on the rich, gene expression-relevant information inherent in the fusion token, which synthesizes data from the target, neighbor, and global tokens. By transferring knowledge from this fusion token to other individual tokens, we significantly enhance the model's predictive accuracy for gene expression levels.
In practice, fully-connected layers, tasked with predicting gene expression levels, are attached to each of the target, neighbor, and global tokens, with average pooling applied beforehand to the target and neighbor tokens. The optimization process involves two key components: 1) minimizing Mean Squared Error (MSE) loss between individual predictors’ outputs and ground-truth values, 2) reducing the MSE loss between each predictor's output and the 'soft target,' which are the predictions derived from the fusion token.

The loss for a given $j_{th}$ token (target, neighbor, or global) is computed as follows:
\begin{equation}
L^{j} = (1-\alpha) \frac{1}{m} \sum^{m}_{k=1} \left\| q^{j}_{k} - y_{k}  \right\|^{2}_{2} + \\
\alpha \frac{1}{m} \sum^{m}_{k=1} \left\| q^{j}_{k} - q_{k}^{F}  \right\|^{2}_{2},
\end{equation}
where $q^{j}_{k}$ is the prediction for the $k_{th}$ gene by the $j_{th}$ token, $q^{F}{k}$ is the fusion token's prediction for the $k_{th}$ gene, and $\alpha$ is a hyperparameter balancing the two aspects of the loss. \\
For the fusion token, we calculate the MSE loss in relation to the actual labels:
\begin{equation}
L^{F} = \frac{1}{m} \sum^{m}_{k=1} \left\| q^{F}_{k} - y_{k}  \right\|^{2}_{2}
\end{equation} 
Ultimately, we optimize the following object function:
\begin{equation}
L = \sum^{3}_{j} L^{j} + L^{F}.
\end{equation}

\section{Experiments}
\label{sec:Experiments}

\hspace{\parindent} In this section, we outline the specifics of the ST data employed in our model's training, detail our experimental setup and evaluation metrics, and provide implementation details.
For more details on experiment settings, please refer to Section 2 of the supplementary material.

\noindent \textbf{ST dataset}
Spatial Transcriptomics (ST) data is characterized by its compilation of numerous spot images within a single slide, each accompanied by corresponding gene expression values. A typical ST dataset contains several hundred spatially resolved spots, with each spot representing the expression values of around 20,000 genes. Visium, the next iteration of ST data, expands this scope to include thousands of spots, each still characterized by the expressions of a similar number of genes.
For our internal validation, we utilize two breast cancer ST datasets \cite{andersson2020spatial,he2020integrating} and one skin cancer ST dataset \cite{ji2020multimodal}. External validation is conducted using Visium data from three independent breast cancer patients. We refer to the breast cancer ST dataset from \cite{andersson2020spatial} as the BC1 dataset, the one from \cite{he2020integrating} as the BC2 dataset, and the skin cancer dataset from \cite{ji2020multimodal} as the SCC dataset. \\
\noindent \textbf{Experiment Setup and Evaluation Metrics}
To mitigate potential overfitting due to the limited size of our datasets, we employ cross-validation for model performance evaluation across the three ST datasets. Consistent with our earlier mention, we ensure that samples from the same patients are exclusively allocated to either the training or the test dataset, avoiding any overlap.
Specifically, we adopt a leave-one-patient-out cross-validation approach for the BC1 dataset (n sample=36, n patient=8) and the SCC dataset (n sample=12, n patient=4), using samples from a single patient for sequential validation. For the BC2 dataset, which has a larger sample size (n sample=68, n patient=23), we conduct 8-fold cross-validation, with careful consideration to keep samples from the same patient within the same data partition.
To extend our model's evaluation to independent datasets, we use three breast cancer Visium datasets from 10x Genomics, training on the BC1 dataset and testing on each Visium dataset. Our evaluation metrics include the Pearson correlation coefficient (PCC), mean squared error (MSE), and mean absolute error (MAE). PCC is computed for each gene across all spots in a sample, and we report both the mean PCC for all genes (PCC(M)) and the mean PCC for highly predictive genes (PCC(H)). The highly predictive genes are identified through a cross-validation ranking process, where the top 50 genes are determined based on their average rank across all folds. \\
\begin{table*}[ht]
\label{t4}
\footnotesize
\begin{adjustbox}{width=1\textwidth}
\begin{tabular}{c|c|cccc|cccc|cccc}
\noalign{\smallskip}\noalign{\smallskip}\hline\hline
 & & \multicolumn{4}{c|}{10X Visium-1} & \multicolumn{4}{c|}{10X Visium-2} & \multicolumn{4}{c}{10X Visium-3} \\
Source & Model & MSE & MAE & PCC(M) & PCC(H) & MSE & MAE & PCC(M) & PCC(H) & MSE & MAE & PCC(M) & PCC(H) \\
\hline
& ST-Net \cite{he2020integrating} & 0.423 & 0.505 & -0.026 & -0.000 &
0.395 & 0.492 & 0.091 & 0.193 &
0.424 & 0.508 & -0.032 & 0.008 \\
& EGN \cite{yang2023exemplar} & 0.421 & 0.512 & 0.003 & 0.024 &
0.328 & 0.443 & 0.102 & 0.157 &
0.303 & 0.425 & 0.106 & 0.220 \\
\textbf{Local} & BLEEP \cite{xie2023spatially} & 0.367 & 0.470 & 0.106 & 0.221 &
0.289 & 0.406 & 0.104 & 0.260 &
0.298 & 0.415 & 0.114 & 0.229 \\
 & TEM & \textbf{0.339} & \textbf{0.453} & 0.024 & 0.093 &
\textbf{0.278} & \textbf{0.402} & 0.106 & 0.218 &
0.290 & 0.412 & 0.078 & 0.193 \\
& NEM & 0.444 & 0.515 & 0.089 & 0.259 &
0.391 & 0.482 & 0.105 & 0.290 &
0.393 & 0.483 & 0.036 & 0.175 \\
\hline
\textbf{Global} & GEM & 0.392 & 0.494 & 0.132 & \textbf{0.269} &
0.397 & 0.482 & 0.056 & 0.166 &
0.394 & 0.488 & 0.082 & 0.191 \\
\hline
\rowcolor{beaublue}
 \textbf{Multiple} & TRIPLEX & 0.351 & 0.464 & \textbf{0.136} & 0.241 &
0.282 & 0.407 & \textbf{0.155} & \textbf{0.356} &
\textbf{0.285} & \textbf{0.410} & \textbf{0.118} & \textbf{0.282} \\
 \hline
\hline
\end{tabular}
\end{adjustbox}
\caption{ Generalization performance comparison between other models and ours by PCC(M), PCC(H), MSE, and MAE.}
\label{table:table2}
\end{table*} 
\noindent \textbf{Implementation Details}
For preprocessing, we crop each spot image to 224x224 pixels using the center coordinates of each spot. Neighbor views are obtained by capturing a 1120x1120 image centered on the spot, which is then subdivided into 25 equal-sized sub-images. We select 250 genes for each dataset following the criteria in \cite{he2020integrating}. The gene expression values are normalized by dividing by the sum of expressions in each spot, followed by a log transformation.
To mitigate experimental noise, we adopt the smoothing approach from \cite{he2020integrating}, averaging the gene expression values of each spot with those of its adjacent neighbors.
Our model is optimized using the Adam optimizer \cite{kingma2014adam} with an initial learning rate of 0.0001. The learning rate is adjusted dynamically using a Step LR scheduler, with a step size of 50 and a decay rate of 0.9. During training, we use a batch size of 128. In testing, the model's performance is evaluated on all spots in each WSI per batch, and we report the mean of all validation performances.
\subsection{Cross-validation Performance of TRIPLEX}
\hspace{\parindent} We conduct cross-validation using the three ST datasets to assess TRIPLEX's performance in comparison to baseline models. The total counts of spot images in the BC1, BC2, and SCC datasets are 13,620, 68,050, and 23,205, respectively.

\noindent \textbf{Baselines}
Our model's performance was benchmarked against existing models, including 1) local-based models (ST-Net\cite{he2020integrating}, EGN\cite{yang2023exemplar}, BLEEP\cite{xie2023spatially}) and 2) global-based models (HisToGene\cite{pang2021leveraging}, Hist2ST\cite{zeng2022spatial}). For a consistent evaluation, the same ResNet18 used in TRIPLEX was applied as the feature extractor in EGN and the image encoder in BLEEP. We also compared TRIPLEX with simpler models focusing on single information types: Target Encoding Model (TEM), Neighbor Encoding Model (NEM), and Global Encoding Model (GEM). Implementation details for all baseline models are available in Section 2 of the supplementary material.

\noindent \textbf{Result Comparison}
As shown in Table \ref{table:table1}, TRIPLEX outperforms all previous models and demonstrates superior performance over the individual modules within TRIPLEX across most evaluation metrics. Notably, GEM demonstrates substantial effectiveness, underscoring the crucial role of global interactions in accurately predicting gene expression. However, TRIPLEX, which integrates the target and neighbor view information with the global view, yields the most notable improvement. Compared to EGN, one of the best-performing existing models, TRIPLEX achieves substantial increases in PCC(M) and PCC(H) across all datasets. Specifically, in the BC1 dataset, there is an improvement of 0.117 in PCC(M) and 0.169 in PCC(H); in the BC2 dataset, an increase of 0.095 in PCC(M) and 0.149 in PCC(H); and in the SCC dataset, a rise of 0.093 in PCC(M) and 0.102 in PCC(H). Despite integrating three types of information, TRIPLEX maintains a parameter count comparable to other top-performing models (see supplementary table 1). The variances between our results and those reported in the original works of the baseline models can be attributed to differences in 1) cross-validation strategies, 2) normalization methods, and 3) metric calculations. For an in-depth explanation of these differences, refer to Section 3 of the supplementary material.

\subsection{Generalization performance of TRIPLEX}
\hspace{\parindent} For external validation, we preprocess Visium data similarly to the ST data and evaluate model performance on three individual Visium samples.

\noindent \textbf{Result comparison}
In this set of experiments, TRIPLEX is benchmarked against the three best-performing models identified in prior tests. According to Table \ref{table:table2}, TRIPLEX shows robust performance on unseen Visium data, which is formatted differently from the training data. It outperforms existing models (ST-Net, EGN, BLEEP) across MSE, MAE, PCC (M), and PCC (H). Notably, TRIPLEX consistently achieves impressive PCC (H) scores (average 0.293), indicating its potential applicability in clinical settings.

\subsection{Visualization of Cancer Marker Genes}
\begin{figure}[!t]
    \centering
    \includegraphics[width=1\linewidth]{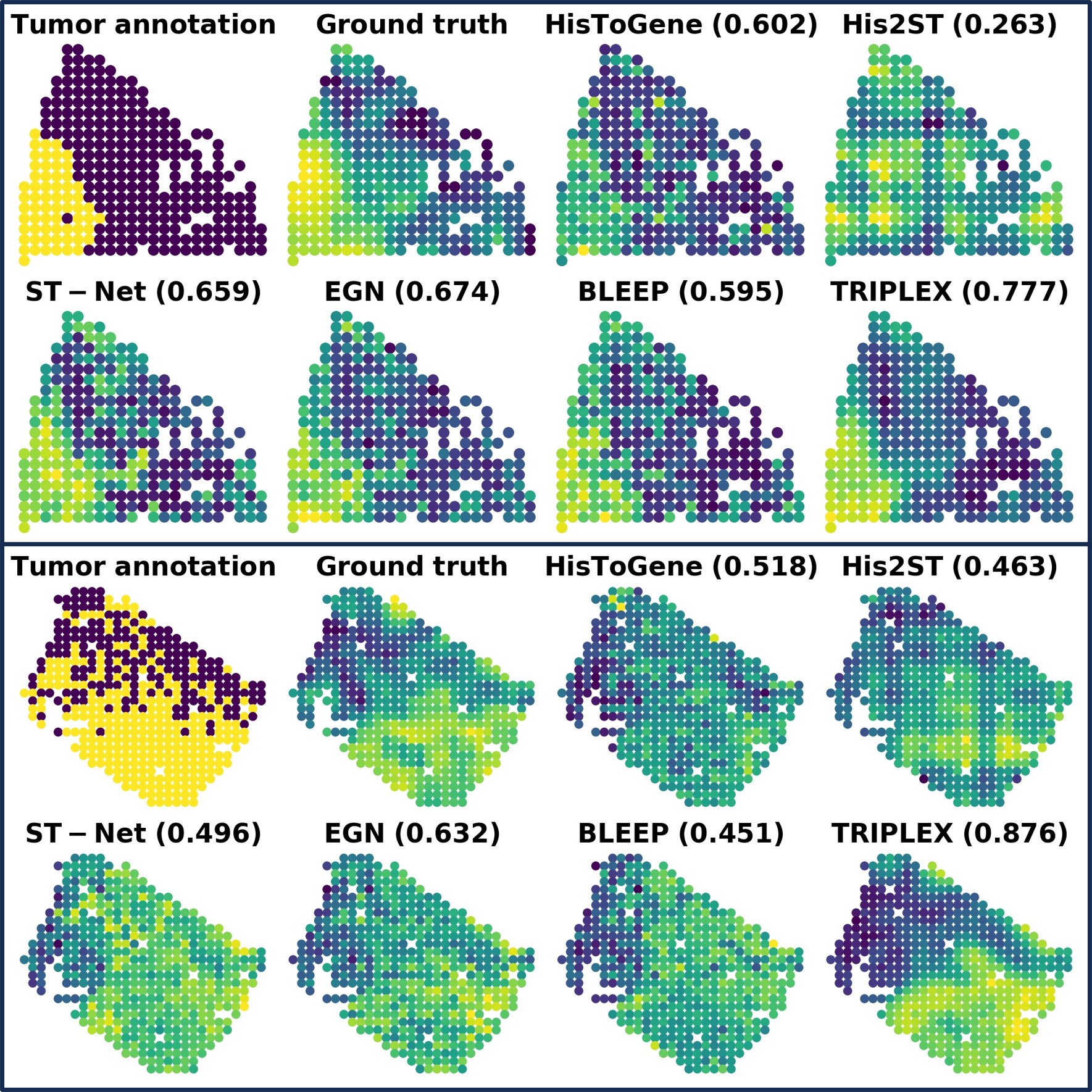}
    \caption{The visualization includes tumor region annotations by pathologists, ground truth for GNAS expression levels, and predicted GNAS expression levels from HisToGene, Hist2ST, ST-Net, EGN, BLEEP, and TRIPLEX, in samples from datasets BC1 and BC2. The PCC between the ground truth and predicted values is displayed for each model.}
    \label{fig:Fig4}
\end{figure}

\hspace{\parindent} Among the highly predictive genes identified by our model are known breast cancer markers such as CLDN4 and GNAS \cite{morin2005claudin,jin2019elevated}, which could aid pathological diagnosis. In the cross-validation for the GNAS gene, TRIPLEX significantly outperforms all existing models in both BC1 and BC2 datasets. For instance, in the BC1 dataset, the PCCs are 0.359 (HisToGene), 0.286 (Hist2ST), 0.411 (ST-Net), 0.374 (EGN), 0.338 (BLEEP), and 0.583 (TRIPLEX). Similarly, in the BC2 dataset, the PCCs are 0.282 (HisToGene), 0.138 (Hist2ST), 0.371 (ST-Net), 0.341 (EGN), 0.0669 (BLEEP), and 0.554 (TRIPLEX).
We further provide visualizations of the predicted values for GNAS alongside its ground truth values for each dataset. As depicted in Figure \ref{fig:Fig4}, we notice that TRIPLEX not only exhibits a higher PCC with the actual ground truth values but also demonstrates a greater visual congruence with the actual gene expression patterns. Consequently, it appears to be more effective in assisting pathologists in clinical diagnostics. More visualizations are available in the supplementary material.

\subsection{Ablation study}

\hspace{\parindent} In this section, we demonstrate the contributions of each method to gene expression prediction through an ablation study of our proposed model. Specifically, we aim to observe the contributions of the three modules in our model and investigate the extent to which the position encoding generator and our proposed fusion approach contribute to gene expression prediction. Here we only present experimental results on the SCC dataset, but we have observed similar outcomes on other datasets as well. Additional experiment results can be found in supplementary material.

\noindent \textbf{Individual Modules}
We assess how the fusion of information from TEM, NEM, and GEM enhances gene expression prediction accuracy. According to Table \ref{table:table3}, it is clear that incorporating features from all three modules achieves the best performance.
Notably, the exclusion of the NEM module results in a substantial decrease in PCC(M), highlighting the critical influence of neighboring interactions in gene expression prediction in TRIPLEX. 
While the absence of the GEM module does not significantly impact performance metrics such as MSE and MAE in this specific dataset, ablation studies conducted on alternative datasets reveal the significant contribution of global interactions to the accuracy of gene expression level predictions.
\begin{table}[ht]
\begin{adjustbox}{width=0.48\textwidth}
\begin{tabular}{c|cccc}
\noalign{\smallskip}\noalign{\smallskip}\hline\hline
    & MSE & MAE & PCC(M) & PCC(T) \\
\hline
 w/o TEM & 0.289 & 0.419 & 0.352 & 0.471 \\    
 w/o NEM & 0.271 & 0.408 & 0.330 & 0.439 \\
 w/o GEM & \textbf{0.263} & \textbf{0.402} & 0.358 & 0.481 \\
 TRIPLEX & 0.268 & 0.404 & \textbf{0.374} & \textbf{0.490} \\
\hline
\hline
\end{tabular}
\end{adjustbox}
\caption{Ablation study for the individual modules}
\label{table:table3}
\end{table} \\

\noindent \textbf{Position Encoding Generator (PEG)} 
We further compare our APEG to different methods of positional encoding, including (1) without PEG and (2) with PEG \cite{chu2021conditional}, as shown in Table \ref{table:table4}. In the conventional PEG implementation, we add zero padding to the global token to make the number of tokens a perfect square, followed by the standard squaring procedure. The results indicate that APEG not only outperforms the model without any PEG but also surpasses the conventional PEG. This implies that APEG more effectively encodes spatial distribution within the global tokens, enhancing the overall model performance.
\begin{table}[ht]
\begin{adjustbox}{width=0.48\textwidth}
\begin{tabular}{c|cccc}
\noalign{\smallskip}\noalign{\smallskip}\hline\hline
    & MSE & MAE & PCC(M) & PCC(T) \\
\hline
 w/o PEG & 0.280 & 0.413 & 0.360 & 0.480 \\
 PEG \cite{chu2021conditional} & 0.276 & 0.411 & 0.364 & 0.479 \\
 APEG & \textbf{0.268} & \textbf{0.404} & \textbf{0.374} & \textbf{0.490} \\
\hline
\hline
\end{tabular}
\end{adjustbox}
\caption{Ablation study PEG}
\label{table:table4}
\end{table} \\
\noindent \textbf{Fusion Method and Fusion Loss}
We also investigate the impact of different fusion methods for multi-resolution features, comparing our model's fusion layer with traditional feature fusion techniques like summation, concatenation, and attentional pooling \cite{ilse2018attention}. Additionally, we assess the contribution of fusion loss to the model's performance. As shown in Table \ref{table:table5}, integrating multi-resolution features using our fusion layer yields superior results compared to these conventional techniques. Moreover, the marked improvement in performance when incorporating fusion loss indicates its effectiveness in integrating the three types of features. This result highlights the significance of our fusion approach in achieving high accuracy in gene expression prediction.
\begin{table}[ht]
\begin{adjustbox}{width=0.48\textwidth}
\begin{tabular}{c|cccc}
\noalign{\smallskip}\noalign{\smallskip}\hline\hline
 Fusion method & MSE & MAE & PCC(M) & PCC(T) \\
\hline
 Summation & 0.297 & 0.425 & 0.341 & 0.464 \ \\    
 Concatenation & 0.293 & 0.422 & 0.348 & 0.474 \\
Attentional pooling & 0.293 & 0.423 & 0.353 & 0.473 \\
fusion layer & \textbf{0.268} & \textbf{0.404} & \textbf{0.374} & \textbf{0.490} \\
\hline
 Fusion loss & MSE & MAE & PCC(M) & PCC(T) \\
\hline
 w/o fusion loss & 0.292 & 0.423 & 0.358 & 0.469 \\    
 w/ fusion loss & \textbf{0.268} & \textbf{0.404} & \textbf{0.374} & \textbf{0.490} \\
\hline
\hline
\end{tabular}
\end{adjustbox}
\caption{Ablation studies for fusion method and fusion loss}
\label{table:table5}
\end{table}


\section{Conclusion}
\label{sec:Conclusion}

\hspace{\parindent} We demonstrate a novel approach for predicting spatial gene expression patterns from WSIs. By incorporating multiple sources of information utilizing various types of transformer and our proposed fusion method, TRIPLEX achieves superior performance compared to all existing approaches in both internal and external evaluations. TRIPLEX has the potential to improve the accuracy and robustness of the predictions for spatial gene expression distribution, paving the way for new discoveries at the interface of WSIs and sequencing. \\
\textbf{Acknowledgments} 
This work was partly supported by Institute of Information \& communications Technology Planning \& Evaluation(IITP) grant funded by the Korea government(MSIT) (No.2019-0-00421, AI Graduate School Support Program(Sungkyunkwan University)) and the Samsung Research Funding \& Incubation Center of Samsung Electronics under Project SRFC-MA2102-05. This work is a study partly supported by domestic scholarships funded by the Kwanjeong Educational Foundation (KEF1464).

{
    \small
    \bibliographystyle{ieeenat_fullname}
    \bibliography{main}

\begin{thebibliography}{29}
\providecommand{\natexlab}[1]{#1}
\providecommand{\url}[1]{\texttt{#1}}
\expandafter\ifx\csname urlstyle\endcsname\relax
  \providecommand{\doi}[1]{doi: #1}\else
  \providecommand{\doi}{doi: \begingroup \urlstyle{rm}\Url}\fi

\bibitem[Andersson et~al.(2020)Andersson, Larsson, Stenbeck, Salm{\'e}n, Ehinger, Wu, Al-Eryani, Roden, Swarbrick, Borg, et~al.]{andersson2020spatial}
Alma Andersson, Ludvig Larsson, Linnea Stenbeck, Fredrik Salm{\'e}n, Anna Ehinger, Sunny Wu, Ghamdan Al-Eryani, Daniel Roden, Alex Swarbrick, {\AA}ke Borg, et~al.
\newblock Spatial deconvolution of her2-positive breast tumors reveals novel intercellular relationships.
\newblock \emph{bioRxiv}, 2020.

\bibitem[Biewald(2020)]{wandb}
Lukas Biewald.
\newblock Experiment tracking with weights and biases, 2020.
\newblock Software available from wandb.com.

\bibitem[Campanella et~al.(2019)Campanella, Hanna, Geneslaw, Miraflor, Werneck Krauss~Silva, Busam, Brogi, Reuter, Klimstra, and Fuchs]{campanella2019clinical}
Gabriele Campanella, Matthew~G Hanna, Luke Geneslaw, Allen Miraflor, Vitor Werneck Krauss~Silva, Klaus~J Busam, Edi Brogi, Victor~E Reuter, David~S Klimstra, and Thomas~J Fuchs.
\newblock Clinical-grade computational pathology using weakly supervised deep learning on whole slide images.
\newblock \emph{Nature medicine}, 25\penalty0 (8):\penalty0 1301--1309, 2019.

\bibitem[Chen et~al.(2022)Chen, Chen, Li, Chen, Trister, Krishnan, and Mahmood]{chen2022scaling}
Richard~J Chen, Chengkuan Chen, Yicong Li, Tiffany~Y Chen, Andrew~D Trister, Rahul~G Krishnan, and Faisal Mahmood.
\newblock Scaling vision transformers to gigapixel images via hierarchical self-supervised learning.
\newblock In \emph{Proceedings of the IEEE/CVF Conference on Computer Vision and Pattern Recognition}, pages 16144--16155, 2022.

\bibitem[Chu et~al.(2021)Chu, Tian, Zhang, Wang, Wei, Xia, and Shen]{chu2021conditional}
Xiangxiang Chu, Zhi Tian, Bo Zhang, Xinlong Wang, Xiaolin Wei, Huaxia Xia, and Chunhua Shen.
\newblock Conditional positional encodings for vision transformers.
\newblock \emph{arXiv preprint arXiv:2102.10882}, 2021.

\bibitem[Ciga et~al.(2022)Ciga, Xu, and Martel]{ciga2022self}
Ozan Ciga, Tony Xu, and Anne~Louise Martel.
\newblock Self supervised contrastive learning for digital histopathology.
\newblock \emph{Machine Learning with Applications}, 7:\penalty0 100198, 2022.

\bibitem[Dosovitskiy et~al.(2020)Dosovitskiy, Beyer, Kolesnikov, Weissenborn, Zhai, Unterthiner, Dehghani, Minderer, Heigold, Gelly, et~al.]{dosovitskiy2020image}
Alexey Dosovitskiy, Lucas Beyer, Alexander Kolesnikov, Dirk Weissenborn, Xiaohua Zhai, Thomas Unterthiner, Mostafa Dehghani, Matthias Minderer, Georg Heigold, Sylvain Gelly, et~al.
\newblock An image is worth 16x16 words: Transformers for image recognition at scale.
\newblock \emph{arXiv preprint arXiv:2010.11929}, 2020.

\bibitem[He et~al.(2020)He, Bergenstr{\aa}hle, Stenbeck, Abid, Andersson, Borg, Maaskola, Lundeberg, and Zou]{he2020integrating}
Bryan He, Ludvig Bergenstr{\aa}hle, Linnea Stenbeck, Abubakar Abid, Alma Andersson, {\AA}ke Borg, Jonas Maaskola, Joakim Lundeberg, and James Zou.
\newblock Integrating spatial gene expression and breast tumour morphology via deep learning.
\newblock \emph{Nature biomedical engineering}, 4\penalty0 (8):\penalty0 827--834, 2020.

\bibitem[Huang et~al.(2017)Huang, Liu, Van Der~Maaten, and Weinberger]{huang2017densely}
Gao Huang, Zhuang Liu, Laurens Van Der~Maaten, and Kilian~Q Weinberger.
\newblock Densely connected convolutional networks.
\newblock In \emph{Proceedings of the IEEE conference on computer vision and pattern recognition}, pages 4700--4708, 2017.

\bibitem[Ilse et~al.(2018)Ilse, Tomczak, and Welling]{ilse2018attention}
Maximilian Ilse, Jakub Tomczak, and Max Welling.
\newblock Attention-based deep multiple instance learning.
\newblock In \emph{International conference on machine learning}, pages 2127--2136. PMLR, 2018.

\bibitem[Ji et~al.(2020)Ji, Rubin, Thrane, Jiang, Reynolds, Meyers, Guo, George, Mollbrink, Bergenstr{\aa}hle, et~al.]{ji2020multimodal}
Andrew~L Ji, Adam~J Rubin, Kim Thrane, Sizun Jiang, David~L Reynolds, Robin~M Meyers, Margaret~G Guo, Benson~M George, Annelie Mollbrink, Joseph Bergenstr{\aa}hle, et~al.
\newblock Multimodal analysis of composition and spatial architecture in human squamous cell carcinoma.
\newblock \emph{Cell}, 182\penalty0 (2):\penalty0 497--514, 2020.

\bibitem[Jin et~al.(2019)Jin, Zhu, Cui, Tang, Xie, and Ren]{jin2019elevated}
X Jin, L Zhu, Z Cui, J Tang, M Xie, and G Ren.
\newblock Elevated expression of gnas promotes breast cancer cell proliferation and migration via the pi3k/akt/snail1/e-cadherin axis.
\newblock \emph{Clinical and Translational Oncology}, 21:\penalty0 1207--1219, 2019.

\bibitem[Kanavati et~al.(2020)Kanavati, Toyokawa, Momosaki, Rambeau, Kozuma, Shoji, Yamazaki, Takeo, Iizuka, and Tsuneki]{kanavati2020weakly}
Fahdi Kanavati, Gouji Toyokawa, Seiya Momosaki, Michael Rambeau, Yuka Kozuma, Fumihiro Shoji, Koji Yamazaki, Sadanori Takeo, Osamu Iizuka, and Masayuki Tsuneki.
\newblock Weakly-supervised learning for lung carcinoma classification using deep learning.
\newblock \emph{Scientific reports}, 10\penalty0 (1):\penalty0 1--11, 2020.

\bibitem[Kingma and Ba(2014)]{kingma2014adam}
Diederik~P Kingma and Jimmy Ba.
\newblock Adam: A method for stochastic optimization.
\newblock \emph{arXiv preprint arXiv:1412.6980}, 2014.

\bibitem[Kipf and Welling(2016)]{kipf2016semi}
Thomas~N Kipf and Max Welling.
\newblock Semi-supervised classification with graph convolutional networks.
\newblock \emph{arXiv preprint arXiv:1609.02907}, 2016.

\bibitem[Korsunsky et~al.(2019)Korsunsky, Millard, Fan, Slowikowski, Zhang, Wei, Baglaenko, Brenner, Loh, and Raychaudhuri]{korsunsky2019fast}
Ilya Korsunsky, Nghia Millard, Jean Fan, Kamil Slowikowski, Fan Zhang, Kevin Wei, Yuriy Baglaenko, Michael Brenner, Po-ru Loh, and Soumya Raychaudhuri.
\newblock Fast, sensitive and accurate integration of single-cell data with harmony.
\newblock \emph{Nature methods}, 16\penalty0 (12):\penalty0 1289--1296, 2019.

\bibitem[Li et~al.(2021)Li, Li, and Eliceiri]{li2021dual}
Bin Li, Yin Li, and Kevin~W Eliceiri.
\newblock Dual-stream multiple instance learning network for whole slide image classification with self-supervised contrastive learning.
\newblock In \emph{Proceedings of the IEEE/CVF conference on computer vision and pattern recognition}, pages 14318--14328, 2021.

\bibitem[Lu et~al.(2021)Lu, Williamson, Chen, Chen, Barbieri, and Mahmood]{lu2021data}
Ming~Y Lu, Drew~FK Williamson, Tiffany~Y Chen, Richard~J Chen, Matteo Barbieri, and Faisal Mahmood.
\newblock Data-efficient and weakly supervised computational pathology on whole-slide images.
\newblock \emph{Nature biomedical engineering}, 5\penalty0 (6):\penalty0 555--570, 2021.

\bibitem[Morin(2005)]{morin2005claudin}
Patrice~J Morin.
\newblock Claudin proteins in human cancer: promising new targets for diagnosis and therapy.
\newblock \emph{Cancer research}, 65\penalty0 (21):\penalty0 9603--9606, 2005.

\bibitem[Pang et~al.(2021)Pang, Su, and Li]{pang2021leveraging}
Minxing Pang, Kenong Su, and Mingyao Li.
\newblock Leveraging information in spatial transcriptomics to predict super-resolution gene expression from histology images in tumors.
\newblock \emph{bioRxiv}, 2021.

\bibitem[Radford et~al.(2021)Radford, Kim, Hallacy, Ramesh, Goh, Agarwal, Sastry, Askell, Mishkin, Clark, et~al.]{radford2021learning}
Alec Radford, Jong~Wook Kim, Chris Hallacy, Aditya Ramesh, Gabriel Goh, Sandhini Agarwal, Girish Sastry, Amanda Askell, Pamela Mishkin, Jack Clark, et~al.
\newblock Learning transferable visual models from natural language supervision.
\newblock In \emph{International conference on machine learning}, pages 8748--8763. PMLR, 2021.

\bibitem[Shao et~al.(2021)Shao, Bian, Chen, Wang, Zhang, Ji, et~al.]{shao2021transmil}
Zhuchen Shao, Hao Bian, Yang Chen, Yifeng Wang, Jian Zhang, Xiangyang Ji, et~al.
\newblock Transmil: Transformer based correlated multiple instance learning for whole slide image classification.
\newblock \emph{Advances in Neural Information Processing Systems}, 34:\penalty0 2136--2147, 2021.

\bibitem[Shaw et~al.(2018)Shaw, Uszkoreit, and Vaswani]{shaw2018self}
Peter Shaw, Jakob Uszkoreit, and Ashish Vaswani.
\newblock Self-attention with relative position representations.
\newblock \emph{arXiv preprint arXiv:1803.02155}, 2018.

\bibitem[St{\aa}hl et~al.(2016)St{\aa}hl, Salm{\'e}n, Vickovic, Lundmark, Navarro, Magnusson, Giacomello, Asp, Westholm, Huss, et~al.]{staahl2016visualization}
Patrik~L St{\aa}hl, Fredrik Salm{\'e}n, Sanja Vickovic, Anna Lundmark, Jos{\'e}~Fern{\'a}ndez Navarro, Jens Magnusson, Stefania Giacomello, Michaela Asp, Jakub~O Westholm, Mikael Huss, et~al.
\newblock Visualization and analysis of gene expression in tissue sections by spatial transcriptomics.
\newblock \emph{Science}, 353\penalty0 (6294):\penalty0 78--82, 2016.

\bibitem[Trockman and Kolter(2022)]{trockman2022patches}
Asher Trockman and J~Zico Kolter.
\newblock Patches are all you need?
\newblock \emph{arXiv preprint arXiv:2201.09792}, 2022.

\bibitem[Xie et~al.(2023)Xie, Pang, Bader, and Wang]{xie2023spatially}
Ronald Xie, Kuan Pang, Gary~D. Bader, and Bo Wang.
\newblock Spatially resolved gene expression prediction from h\&e histology images via bi-modal contrastive learning, 2023.

\bibitem[Yang et~al.(2022)Yang, Hossain, Gedeon, and Rahman]{yang2022s2fgan}
Yan Yang, Md~Zakir Hossain, Tom Gedeon, and Shafin Rahman.
\newblock S2fgan: semantically aware interactive sketch-to-face translation.
\newblock In \emph{Proceedings of the IEEE/CVF Winter Conference on Applications of Computer Vision}, pages 1269--1278, 2022.

\bibitem[Yang et~al.(2023)Yang, Hossain, Stone, and Rahman]{yang2023exemplar}
Yan Yang, Md~Zakir Hossain, Eric~A Stone, and Shafin Rahman.
\newblock Exemplar guided deep neural network for spatial transcriptomics analysis of gene expression prediction.
\newblock In \emph{Proceedings of the IEEE/CVF Winter Conference on Applications of Computer Vision}, pages 5039--5048, 2023.

\bibitem[Zeng et~al.(2022)Zeng, Wei, Yu, Yin, Yuan, Li, Tang, Lu, and Yang]{zeng2022spatial}
Yuansong Zeng, Zhuoyi Wei, Weijiang Yu, Rui Yin, Yuchen Yuan, Bingling Li, Zhonghui Tang, Yutong Lu, and Yuedong Yang.
\newblock Spatial transcriptomics prediction from histology jointly through transformer and graph neural networks.
\newblock \emph{Briefings in Bioinformatics}, 23\penalty0 (5):\penalty0 bbac297, 2022.

\end{thebibliography}
}

\setcounter{page}{1}
\maketitlesupplementary
\renewcommand\thesection{\Alph{section}}

\section{Evaluation methodology for multi-output regression problem}
In this study, our objective is to address the multi-output regression problem in \textit{n} distinct spots within a Whole Slide Image (WSI). We formally define this problem as:
\[
f: X \rightarrow Y \in \mathbb{R}^{n \times m}
\]
where \textit{X} represents the set of \textit{n} input images, and \textit{Y} denotes the expression levels of \textit{m} genes across these \textit{n} spots. To tackle this problem, we employ three evaluation metrics: the Pearson Correlation Coefficient (PCC), Mean Squared Error (MSE), and Mean Absolute Error (MAE). Our choice of these metrics is grounded in their distinct advantages. Firstly, the PCC offers insights into the linear relationship between predicted and actual target values, both in strength and direction. Secondly, the MSE is a robust measure of the average squared discrepancies between predicted and actual targets, reflecting the model's accuracy and sensitivity to errors. Lastly, the MAE provides an interpretable measure of the average absolute differences between predictions and actual targets, advantageous for its lower sensitivity to outliers compared to MSE.

We assess each model's performance on a per-slide basis. For the $j_{th}$ gene, the PCC, MAE, and MSE are calculated as follows:

\[
PCC_j = \frac{\sum_{i=1}^n (\hat{y}_{i,j} - \bar{\hat{y}}_{\cdot,j})(y_{i,j} - \bar{y}_{\cdot,j})}{\sqrt{\sum_{i=1}^n (\hat{y}_{i,j} - \bar{\hat{y}}_{\cdot,j})^2}\sqrt{\sum_{i=1}^n (y_{i,j} - \bar{y}_{\cdot,j})^2}}
\]
\[
\operatorname{MAE}(Y, \hat{Y}) = \frac{1}{n \times m} \sum_{i=1}^n \sum_{j=1}^m |y_{i,j} - \hat{y}_{i,j}|
\]
\[
\operatorname{MSE}(Y, \hat{Y}) = \frac{1}{n \times m} \sum_{i=1}^n \sum_{j=1}^m (y_{i,j} - \hat{y}_{i,j})^2
\]
Here, $\hat{y}_{i,j}$ represents the predicted expression level of the $j_{th}$ gene in the $i_{th}$ spot, and $m$ and $n$ are numbers of genes to be predicted and spots in a WSI, respectively. For PCC, the average value across $m$ number of genes is computed as:
\[
PCC = \frac{1}{m} \sum_{j=1}^{m} PCC_j
\] 
When multiple slides are involved in testing, we calculate PCC, MAE, and MSE for each slide using the above methodology and then report the average values across all slides.

\section{Method Details}

\subsection{Processing of input data}
\noindent \textbf{Target spot image}
For the extraction of target spot images, we employ pre-defined center coordinates to obtain images of dimensions 224x224. Subsequently, these images undergo a normalization process where pixel values are adjusted to fall within the range of 0 to 1, prior to their input into the model. During the training phase, we enhance the robustness of our model by applying image augmentation techniques. These techniques encompass random horizontal and vertical flips, and random rotations of the input images by 90 degrees.

\begin{figure*}[!t]
    \centering
    \includegraphics[width=1\linewidth]{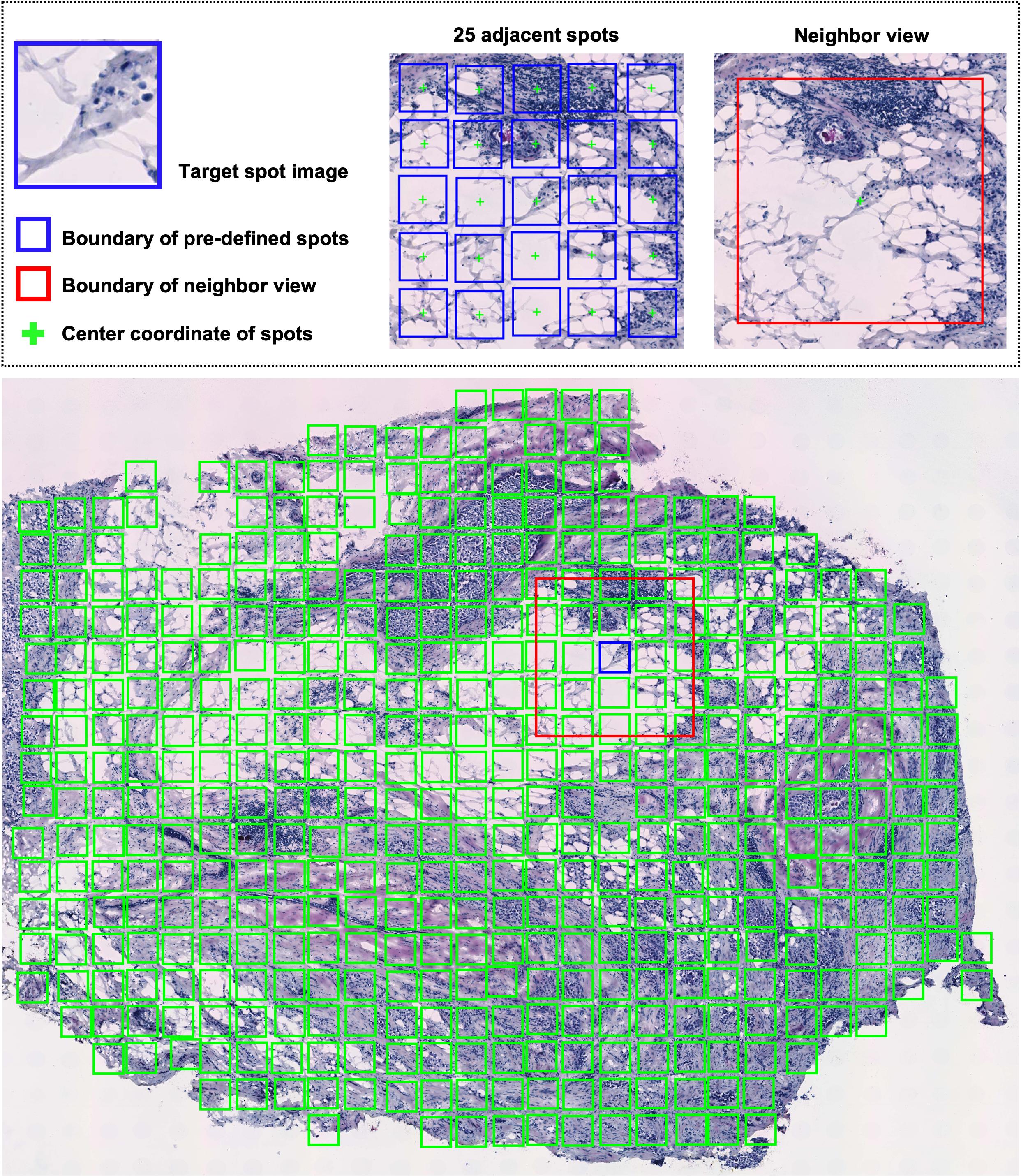}
    \caption{An example of input data for TRIPLEX from BC1 dataset. (Top) Difference between the input data used in NEM and the 25 adjacent spot images around the target spot image. The pre-defined spot image is marked with a blue boundary, while the input data for the NEM model is marked with a red boundary. The '+' within each image indicates the center coordinates. (Bottom) All input data for the same sample. The input data for TEM is marked with a blue boundary, the input data for NEM is marked with a red boundary, and the input data for GEM is marked with a green boundary. (The spot marked with the blue boundary is the target spot image.)}
    \label{fig:sFigure1}
\end{figure*}

\noindent \textbf{Neighbor view}
In processing images sized 1,120x1,120, we commence by segmenting a centrally located 1,120x1,120 patch from the target spot image into 25 uniform sub-patches, each measuring 224x224. It is important to note that these sub-patches differ from the 25 spot images nearest to the target spot, a distinction necessitated by the non-uniform alignment of center coordinates in ST data and the observable gaps between spot images, as depicted in Figure \ref{fig:sFigure1}.

Feature extraction is carried out using a ResNet18 model that has been pre-trained. The significant dissimilarity between histology images and conventional image types presents a challenge, as models trained on datasets like ImageNet might not be directly suitable for WSI analysis. To address this, we utilize a version of ResNet18 that has undergone training on an integrated, multi-organ dataset through self-supervised learning. This training strategy ensures the extraction of features that are robust to variations in staining and resolution, as detailed in \cite{ciga2022self}. The extracted features are then employed as inputs for the neighbor encoder.

\noindent \textbf{Global view}
We engage all spot images contained within a WSI. It's important to clarify that the aggregation of all spot images does not represent the full extent of the WSI. Nonetheless, this comprehensive inclusion of spot images enables us to effectively map the interconnections between these images and approximate the spatial information, as illustrated in Figure \ref{fig:sFigure1}.

For feature extraction, we apply the same methodology used in the neighbor view. This involves cropping spot images to a uniform size of 224x224 and processing them through the pre-trained ResNet18 model. The features extracted from this process are then channeled as inputs into the global encoder for further processes.

\noindent \textbf{Visium data for external test}
We evaluate our model on spatial gene expression data from breast cancer tissues, sourced from 10x Genomics. The datasets employed are as follows:
\begin{itemize}
\item 10X Visium-1: Breast cancer tissue from human (v1), Spatial Gene Expression Dataset by Space Ranger v1.3.0 (2022, Jul 02).
\item 10X Visium-2: Breast cancer tissue from human (v1, Section 1), Spatial Gene Expression Dataset by Space Ranger v1.1.0 (2020, Jun 12).
\item 10X Visium-3: Breast cancer tissue from human (v1, Section 2), Spatial Gene Expression Dataset by Space Ranger v1.1.0 (2020, Jun 12).
\end{itemize}
This dataset represents an enhancement over the ST data used in the training phase, providing high-resolution gene expression profiles with thousands of spots per sample. We apply consistent pre-processing methods across all Visium datasets. The model, initially trained on the BC1 dataset, is subsequently applied to predict 250 genes, initially selected based on their representation in the BC1 dataset.
In cases where any of the 250 genes are not present in a Visium dataset, we exclude those genes from our evaluation. This approach leads to the consistent exclusion of 7 genes across all datasets, a detail we elaborate upon in Figure \ref{fig:sFigure6}.

\subsection{Method details for Target Encoder}
As detailed in the Methods section of the main text, the target encoder embeds target spot images using the pre-trained ResNet18 model \cite{ciga2022self}. This model is fine-tuned to specifically capture fine-grained, target-specific information from the target spot images. In particular, an image with dimensions of 224x224x3 undergoes a transformation into a 7x7x512 feature map after processing through all layers of the ResNet18 model. The resulting features are then reshaped into 49x512 tokens. These tokens are integrated with other tokens -neighbor and global tokens- to form the input for the fusion layer. Concurrently, a separate fully connected layer is linked to the average-pooled token of the target tokens, facilitating independent gene expression prediction. During the training phase, the weights of the target encoder are comprehensively updated to enhance the capture of fine-grained spot information.

\subsection{Method details for Neighbor Encoder}
The neighbor encoder is designed to embed local information surrounding the target spot. It processes features extracted from 25 images, each of size 224x224, representing the neighbor view. This approach closely aligns with the Vision Transformer (ViT) \cite{dosovitskiy2020image} architecture, incorporating self-attention mechanisms with relative position encoding and fully-connected layers applied to the input tokens. A key deviation from the standard ViT model is the exclusion of the patch embedding module for 2D images. Instead, we directly utilize pre-extracted features as our input values. Details regarding specific hyperparameters and their settings will be discussed in the forthcoming section, "Additional Implementation Details and Experimental Results." In a manner similar to the target layer, an additional fully-connected layer is attached to the pooled token of the neighbor tokens. This layer is also specifically tasked with the prediction of gene expression.

\subsection{Method details for Global Encoder}
\begin{figure*}[!t]
    \centering
    \includegraphics[width=0.6 \linewidth]{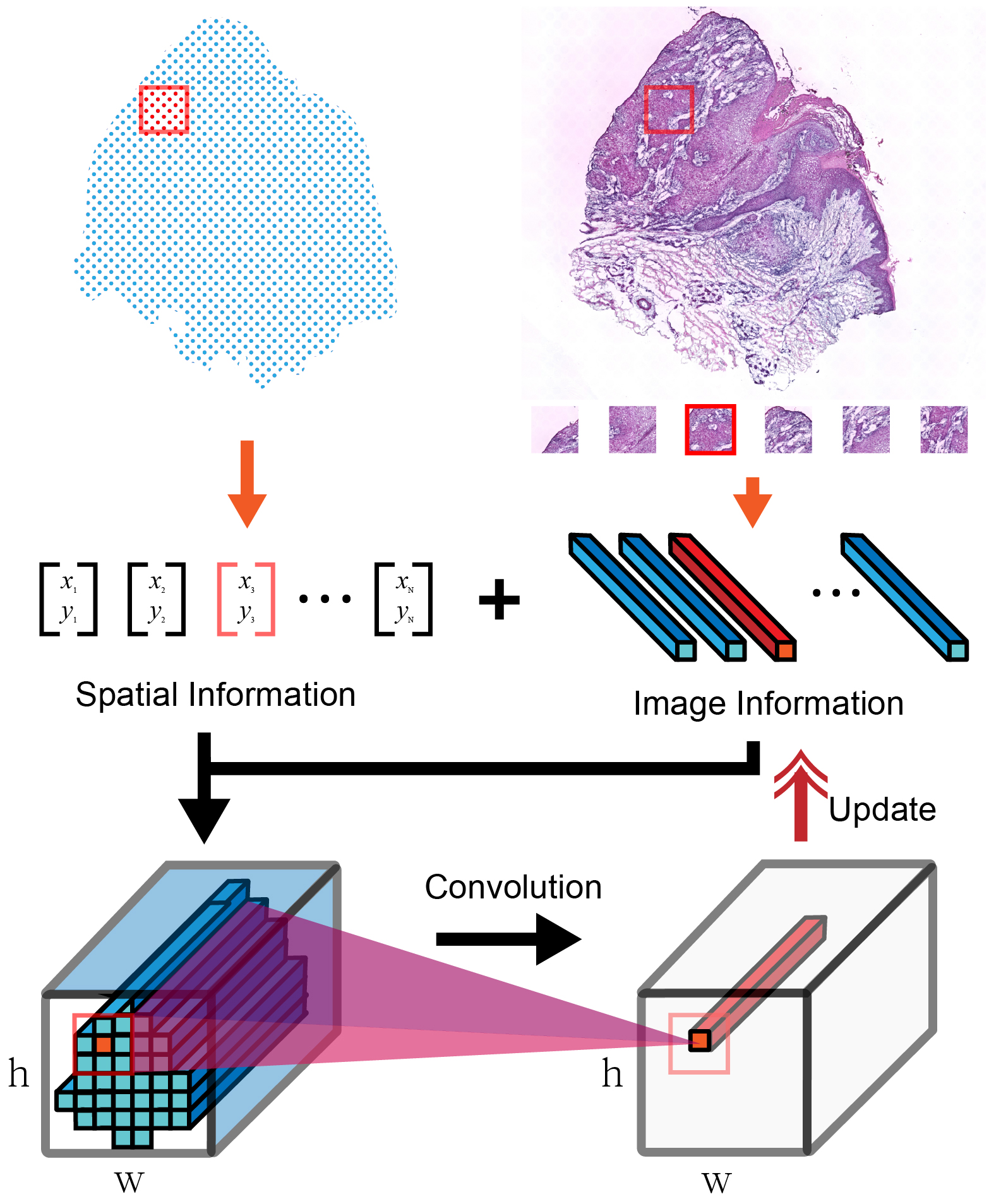}
    \caption{Overview of proposed positional encoding for histology images (APEG). We utilize the coordinates of each spot to reposition the feature token to its original location, apply convolution, and then restore it to its original shape.}
    \label{fig:sFigure2}
\end{figure*}

The global encoder is composed of transformer blocks integrated with the Atypical Position Encoding Generator (APEG). The operational flow of APEG is illustrated in Figure \ref{fig:sFigure2}.

\noindent \textbf{Implementation of APEG}
APEG's process begins with the rearrangement of spot features based on their relative coordinates. This involves constructing a sparse matrix in the Coordinate Format (COO) using Pytorch. In this matrix, indices represent adjusted normalized coordinates, starting from a minimum value of 0, and feature values correspond to the non-zero elements of the matrix. This sparse matrix is subsequently converted into a dense format to facilitate the application of convolutional layers, after which it is restored to its original sparse format.

For the global layer, as opposed to employing a pooling operation, the approach involves retrieving tokens that correspond to each target spot. These tokens are then connected to a fully connected layer, which is tailored to independently predict gene expression.

\subsection{Method details for Fusion Layer}
The fusion layer is specifically designed to integrate information from the global token with corresponding neighbor and target tokens. In this process, the global token actively exchanges information with all other tokens. However, the target and neighbor tokens, which collectively form a larger set of tokens, are structured to avoid direct interaction or information exchange among themselves. In a more detailed mechanism, for each spot, a global token functions as the query, while neighbor and target tokens are utilized as key and value elements. The overall time complexity of the fusion layer is represented as $O(n^{Ta}+n^{Ne})$, where $n^{Ta}$ and $n^{Ne}$ denote the numbers of target and neighbor tokens, respectively.
Following this interactive process, the aggregated tokens are processed through a fully connected layer to yield the final prediction output. This approach is identified as more computationally efficient than applying attention mechanisms across the complete set of tokens, which would result in time complexity of  $O((n^{Ta}+n^{Ne}+1)^2)$. Furthermore, it has shown superior performance compared to traditional feature fusion methods, a claim substantiated by our experimental results.

\subsection{Implementation Details for Baselines}
This subsection details the implementation nuances of our baseline models, ensuring a consistent comparison framework with our proposed TRIPLEX model. In preprocessing the input data for all baseline models, we adhere to the same steps as outlined in Section 4 in the main text, which are also employed in TRIPLEX.

\noindent \textbf{ST-Net}
ST-Net \cite{he2020integrating} utilizes a DenseNet121 \cite{huang2017densely} model, pre-trained on ImageNet, with minimal modifications (only replacing the final output layer) and is fine-tuned on ST data using transfer learning. Our implementation strictly adheres to this scheme.

\noindent \textbf{HisToGene and Hist2ST}
These models employ spot images of size 112x112, as used in their respective studies \cite{pang2021leveraging,zeng2022spatial}. We maintain their overall architectural framework while adapting data normalization and image augmentation techniques to align with our approach. Details on hyperparameter selection can be found in Section 3.6.

\noindent \textbf{EGN}
For EGN \cite{yang2023exemplar}, in addition to following the preprocessing steps of TRIPLEX, we implement the method proposed by \cite{yang2023exemplar} to obtain \textit{k} exemplars for all spots. We replace the unsupervised SF2GAN-based model from \cite{yang2022s2fgan} with a pretrained ResNet18 model used in TRIPLEX. This decision, informed by the results in \cite{yang2023exemplar}, ensures a fair comparison by maintaining consistent feature extraction strategies across models.

\noindent \textbf{BLEEP}
BLEEP \cite{xie2023spatially} is a bi-modal learning model designed to co-embed histology images and gene expression levels. In our implementation, we adhere to the core architecture of BLEEP, with minor hyperparameter adjustments as detailed in Section 3.6. For inference, BLEEP utilizes the top-k nearest gene expression levels to predict gene expression for query spot images. Among the three methods suggested - "simple" (using the top 1 value), "simple average" (averaging the top k values), and "weighted average" (using a weighted average of the top k values) - we employ the "simple average" approach, specifically using the top-50 nearest gene expression levels. This choice was based on its superior performance in our tests. Additionally, we opted not to use Harmony \cite{korsunsky2019fast} for batch correction of gene expression levels, as Harmony is geared towards correcting PCA embeddings rather than raw gene expression levels. Given our dataset's limited slide range (12 to 68), such batch correction might inadvertently reduce the training data's diversity, thereby increasing the risk of overfitting.

\noindent \textbf{TEM,NEM,GEM}
We replicate the three derivative models from TRIPLEX: the Target Encoding Model (TEM), Neighbor Encoding Model (NEM), and Global Encoding Model (GEM). Each model is specialized to process distinct views: TEM utilizes the target spot image, NEM focuses on the neighbor view, and GEM deals with the global view. Their primary objective is to predict gene expression levels based on their respective input data. Consistency with TRIPLEX is maintained in terms of architecture and hyperparameters for these models.

\section{Additional implementation details and experimental results}
\subsection{Description of datasets}
\begin{figure*}[!t]
    \centering
    \includegraphics[width=1\linewidth]{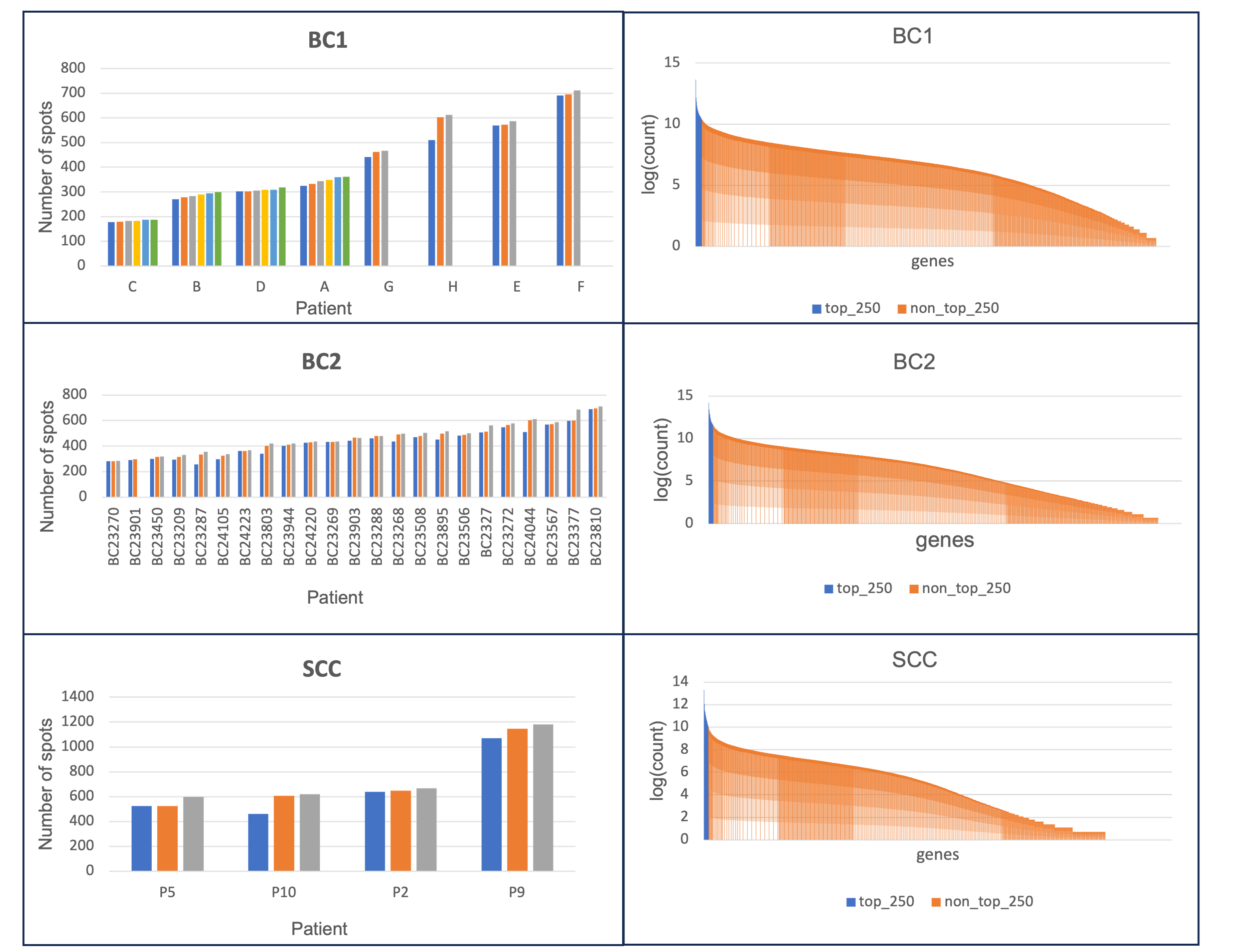}
    \caption{Dataset summary of ST data used for cross-validation. (Left) Number of spots per sample in each dataset. The x-axis label represents each patient, with multiple samples existing for every patient. (Right) Log-transformed count values for each gene in the datasets. The 250 genes utilized in this study correspond to the top genes within the blue region.}
    \label{fig:sFigure3}
\end{figure*}

\begin{figure*}[!t]
    \centering
    \includegraphics[width=1\linewidth]{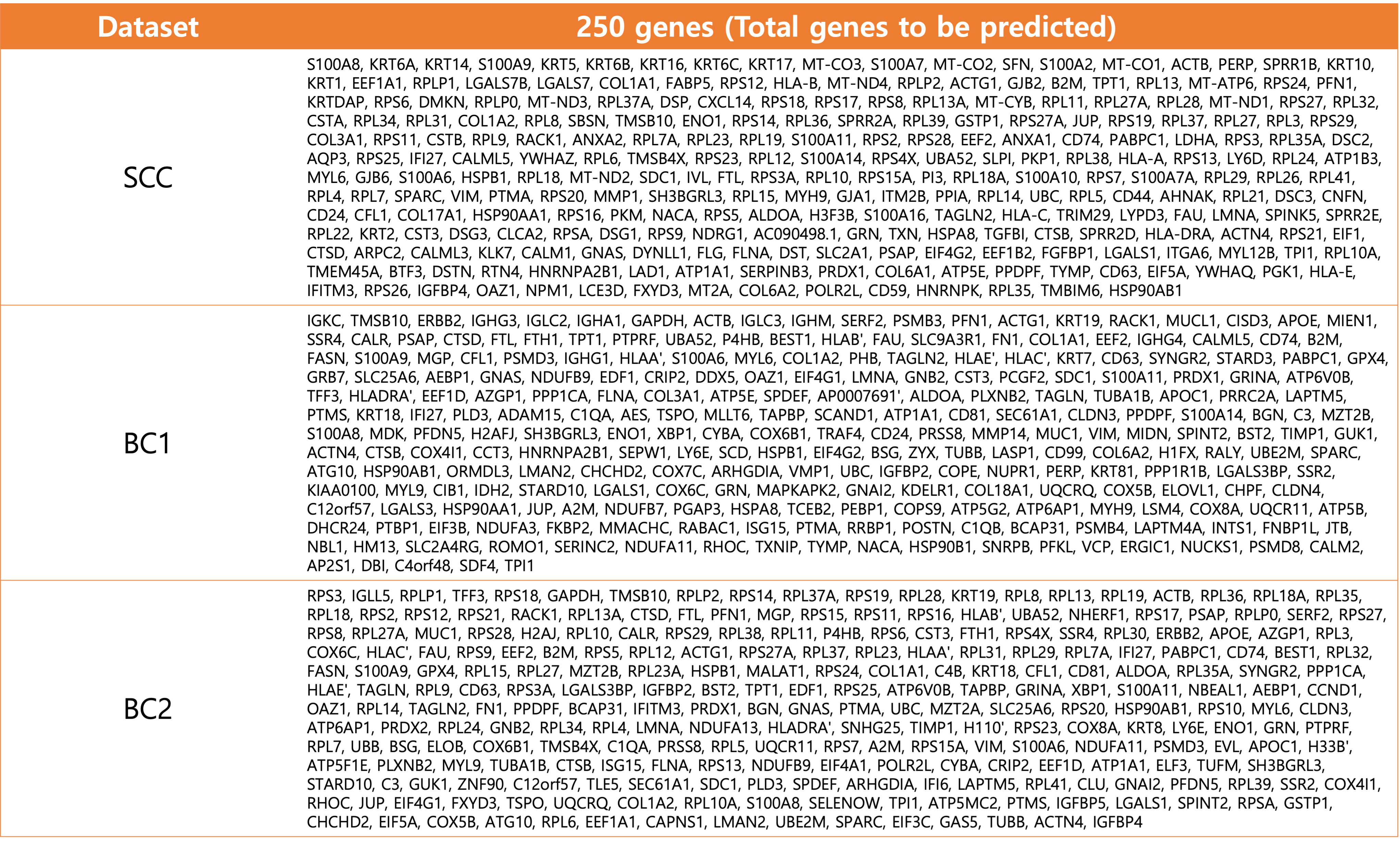}
    \caption{}
    \label{fig:sFigure4}
\end{figure*}

\begin{figure*}[!t]
    \centering
    \includegraphics[width=1\linewidth]{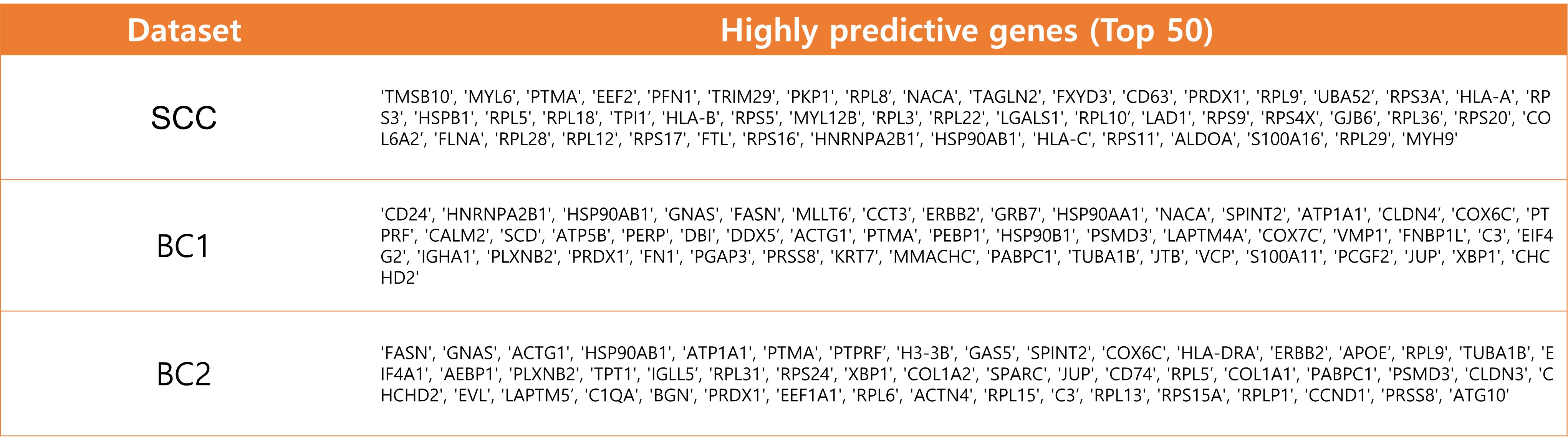}
    \caption{}
    \label{fig:sFigure5}
\end{figure*}

\begin{figure*}[!t]
    \centering
    \includegraphics[width=1\linewidth]{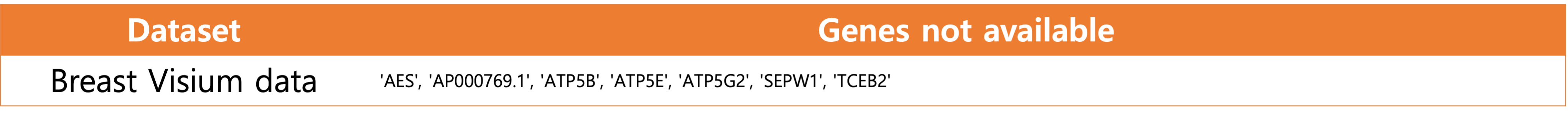}
    \caption{}
    \label{fig:sFigure6}
\end{figure*}

We evaluate our model on three distinct datasets: BC1 and BC2 (breast cancer datasets) and SCC (a skin cancer dataset). We focus on the top 250 genes with high gene expression levels in each dataset as labels for prediction. Furthermore, we calculate the average ranks of well-predicted genes during cross-validation to identify the top 50 "highly predictive genes." These genes are then used to compute the PCC (H). For detailed summaries of each dataset and the specific genes selected, please refer to Figures \ref{fig:sFigure3}, \ref{fig:sFigure4}, and \ref{fig:sFigure5}.

\subsection{Implementation Details for Experiments}
Our approach is implemented using PyTorch (version 1.13.0) and pytorch-lightning (version 1.8.0), and models are trained on a Nvidia RTX A5000 GPU. We employ mixed precision training, utilizing PyTorch native Automatic Mixed Precision (AMP) for efficiency. To ensure reproducibility, the random seed is consistently set at 2021 across all experiments. The training process is capped at a maximum of 200 epochs, with an early stopping mechanism triggered if there is no improvement in the PCC(M) (MSE in case of BLEEP) after 20 epochs.

\subsection{Implementation Details for Ablation Studies}
We assess the impact of omitting individual components and comparing the resulting model performance with the complete TRIPLEX model. Key components of TRIPLEX include: individual modules (TEM, NEM, GEM), each predicting gene expression levels using distinct input data; the Position Encoding Generator (PEG), which infuses positional information into WSIs; and a fusion strategy designed to integrate various types of tokens effectively.

\noindent \textbf{Individual Modules}
In our experimental setup, each module (TEM, NEM, GEM) is individually omitted while maintaining the other components as per the original TRIPLEX configuration. Notably, in scenarios where the GEM is excluded, we introduce a dimensionally equivalent, randomly initialized token in place of the global token. This approach is necessary because, without GEM, there is no medium for information exchange in the fusion layer. 

\noindent \textbf{Position Encoding Generator (PEG)}
We evaluate the significance of our Atypical PEG (APEG), which is engineered to encapsulate positional information within a WSI, on the model's ability to predict gene expression levels. This evaluation involves either removing APEG or substituting it with a traditional PEG as detailed in \cite{chu2021conditional} and Section 4.4 in the main text.

\noindent \textbf{Fusion Method}
To ascertain the efficacy of our proposed fusion layer in amalgamating different token types for the prediction of gene expression levels, we compare TRIPLEX's performance when the fusion layer is replaced with alternative methods: element-wise summation, concatenation, and attentional pooling. In the case of attentional pooling, we dynamically compute feature weights using a neural network, subsequently deriving a weighted sum of the features, as illustrated in \cite{ilse2018attention}.

\subsection{Computational Cost Comparison}
Table \ref{table:stable1} provides a detailed comparison of the computational costs between TRIPLEX and the baseline models, calculated using a single slide sample for each fold in each dataset. This table includes average values for Multiply-Accumulate Operations (MACs), the number of parameters for each model, and both training and testing times across all folds. It's important to note that the training time is gauged based on the duration required to complete 10 epochs.

While TRIPLEX, with its additional inputs, has higher training time compared to other baselines, two observations particularly highlight its efficiency: 1) TRIPLEX's feature extraction technique and integration method efficiently limit its parameters to approximately 20 million, which, while comparable to the baselines, allows TRIPLEX to still achieve state-of-the-art performance. This demonstrates the model's ability to balance complexity with high performance. 2) Additionally, TRIPLEX shows comparable testing times to other leading models (ST-Net, EGN, BLEEP), indicating that its speed remains competitive for practical applications after training. This balance between training complexity and testing efficiency underscores the model's practical applicability in real-world scenarios.

\begin{table}[ht]
\begin{adjustbox}{width=0.48\textwidth}
\begin{tabular}{c|cccc}
\noalign{\smallskip}\noalign{\smallskip}\hline\hline
Dataset & \multicolumn{4}{c}{BC1} \\
    & MACs(G) & \# Param(M) & Training Time(s) & Testing Time(ms) \\
\hline
 ST-Net & 1002 & 7 & 244.2 & 201.7 \\
 HisToGene & 52 & 153 & 291.5 & 2.55 \\
 Hist2ST & 110 & 107 & 254.9 & 7.62 \\
 EGN & 1823 & 162 & 407.6 & 52.67 \\
 BLEEP & 631 & 11 & 119.7 & 109.5 \\
 TRIPLEX & 657 & 22 & 410.9 & 53.21 \\
\hline
Dataset & \multicolumn{4}{c}{BC2} \\
& MACs(G) & \# Param(M) & Training Time(s) & Testing Time(ms) \\
\hline
 ST-Net & 1465 & 7 & 508.7 & 295.0 \\
 HisToGene & 77 & 153 & 329.5 & 2.33 \\
 Hist2ST & 20 & 37 & 193.0 & 5.54 \\
 EGN & 865 & 39 & 722.1 & 31.58 \\
 BLEEP & 923 & 11 & 207.2 & 72.93 \\
 TRIPLEX & 960 & 20 & 1117.3 & 76.27 \\
\hline
Dataset & \multicolumn{4}{c}{SCC} \\
 & MACs(G) & \# Param(M) & Training Time(s) & Testing Time(ms) \\
\hline
 ST-Net & 1928 & 7 & 153.9 & 385.6 \\
 HisToGene & 18 & 27 & 93.70 & 3.12 \\
 Hist2ST & 13 & 14 & 57.5 & 5.37 \\
 EGN & 5563 & 223 & 368.8 & 157.83 \\
 BLEEP & 1215 & 11 & 86.4 & 95.41 \\
 TRIPLEX & 1263 & 19 & 340.3 & 99.90 \\
\hline
\hline
\end{tabular}
\end{adjustbox}
\caption{Computational cost comparison}
\label{table:stable1}
\end{table}

\subsection{Comparison of MAE in the cross-validation experiments}
Table \ref{table:stable2} shows the evaluation of the MAE of the cross-validation results on ST data, which is not included in the main text due to space limitations.

\begin{table}[ht]
\centering
\begin{adjustbox}{width=0.47\textwidth}
\label{t4}
\footnotesize
\begin{tabular}{c|c|c|c|c}
\noalign{\smallskip}\noalign{\smallskip}
\hline
\hline
     &  & \multicolumn{1}{c|}{BC1} & \multicolumn{1}{c|}{BC2} & \multicolumn{1}{c}{SCC} \\
   Source & Model & MAE & MAE & MAE \\
\hline
 & ST-Net \cite{he2020integrating} & $0.389\pm0.03$ & $0.349\pm0.02$ & $0.428\pm0.05$ \\
 & EGN \cite{yang2023exemplar} & $0.377\pm0.04$ & $0.337\pm0.02$ & $0.418\pm0.06$ \\
 \textbf{Local} & BLEEP \cite{xie2023spatially} & $0.401\pm0.03$ & $0.369\pm0.02$ & $0.430\pm0.04$ \\
 & TEM & $0.385\pm0.03$ & $0.336\pm0.02$ & $0.433\pm0.05$ \\
 & NEM & $0.403\pm0.06$ & $0.375\pm0.03$ & $0.481\pm0.10$ \\
\hline
 & HistoGene \cite{pang2021leveraging} & $0.428\pm0.07$ & $0.335\pm0.04$ & $0.415\pm0.07$ \\
 \textbf{Global} & Hist2ST \cite{zeng2022spatial} & $0.413\pm0.07$ & $\textbf{0.333}\pm0.02$ & $0.924\pm0.29$ \\
 & GEM & $0.383\pm0.05$ & $0.352\pm0.02$ & $0.434\pm0.12$ \\
\hline
\rowcolor{beaublue}
\textbf{Multiple}
& TRIPLEX & $\textbf{0.362}\pm0.05$ & $0.343\pm0.02$ & $\textbf{0.404}\pm0.07$ \\
\hline
\hline
\end{tabular}
\end{adjustbox}
\caption{Cross validation result on each ST dataset. The mean and standard deviation of MAE from the cross-validation results are displayed.
}
\label{table:stable2}
\end{table}

\subsection{Contribution of the Neighbor View size}
\begin{figure*}[!t]
    \centering
    \includegraphics[width=1\linewidth]{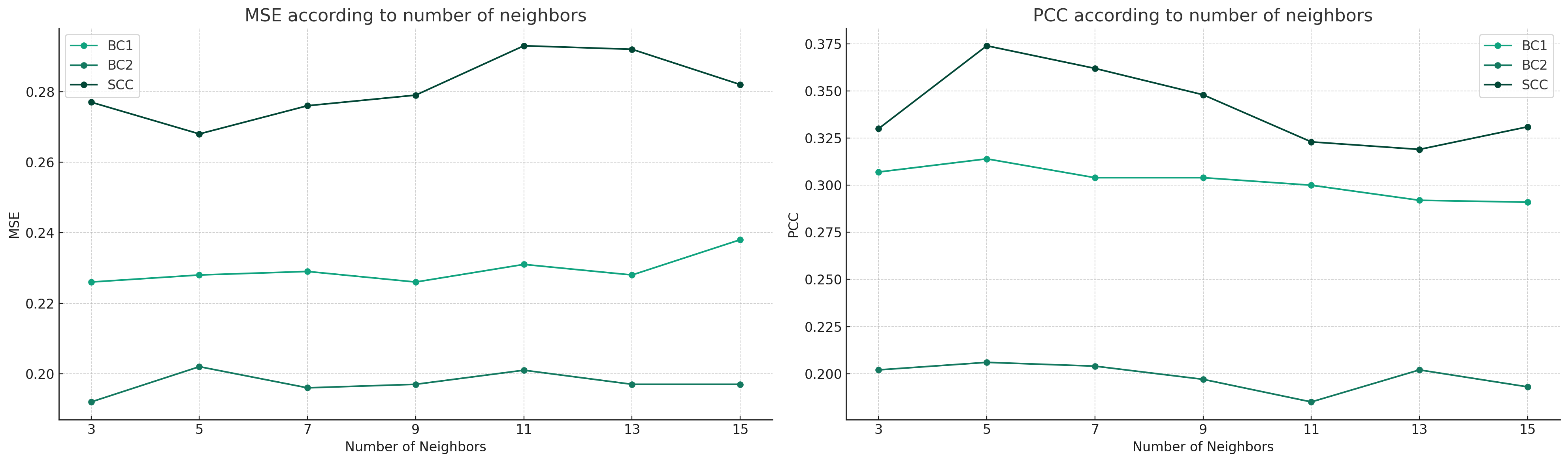}
    \caption{The performance varies with the size of the neighbor view. Variations in MSE (\textbf{Left}) and PCC (M) (\textbf{Right}) relative to the size. "Number of neighbors" represents the count of 224x224 patches along an axis}
    \label{fig:sFigure7}
\end{figure*}

We examine how the performance of TRIPLEX is influenced by the expansion of the neighbor view size. Here, the term "number of neighbors" refers to the count of 224x224 patches along an axis. Results are illustrated in Figure \ref{fig:sFigure7}. When evaluated in terms of MES and PCC, it is observed that enlarging the neighbor view size does not consistently result in performance gains. In fact, as the number of neighbors increases, a decrease in performance is noted. This pattern suggests that 1,120x1,120 sized neighbor view (number of neighbors: 5) is an efficient configuration, striking a balance between capturing detailed neighboring information relevant to the target and maintaining manageable computational costs.

\subsection{Performance Discrepancy Between Our Experimental Results and Existing Implementations}
Our experimental results show notable deviations from those reported in the original publications of the baseline models. We attribute this discrepancy primarily to three factors, as detailed in Section 4.1 in the main text: 1) the use of an alternative cross-validation strategy, 2) a different approach to normalization, and 3) variations in how metrics are calculated.

Specifically, the performance gap observed for HisToGene and Hist2ST can be largely traced back to the first factor. In the original studies, these models are tested on the BC1 and SCC datasets using Leave-one-out-cross-validation (LOOCV), where each sample is treated independently. This approach potentially skews the evaluation, as it allows for the possibility of using replicates from the same sample in both training and testing phases. In contrast, our study employs Leave-one-patient-out-cross-validation (LOPCV), which we believe offers a stricter and more realistic assessment of model performance by ensuring no overlap between training and testing sets for a given patient. Table \ref{table:stable3} compares the results obtained from these two cross-validation methods. As hypothesized, the change to LOPCV significantly affects the performance of both models, reinforcing our assertion about the importance of the rigorous cross-validation approach in model evaluation. 

\begin{table}[ht]
\begin{tabular}{c|cccc}
\noalign{\smallskip}\noalign{\smallskip}\hline\hline
Model & \multicolumn{4}{c}{HisToGene} \\
    & MSE & MAE & PCC(M) & PCC(H) \\
\hline
LOPCV (ours) & 0.314 & 0.428 & 0.168 & 0.302 \\
LOOCV \cite{pang2021leveraging,zeng2022spatial} & 0.223 & 0.364 & 0.186 & 0.315 \\    
 \hline
Model & \multicolumn{4}{c}{HisT2ST} \\
& MSE & MAE & PCC(M) & PCC(H) \\
\hline
LOPCV (ours) & 0.285 & 0.413 & 0.118 & 0.248 \\
LOOCV \cite{pang2021leveraging,zeng2022spatial} & 0.163 & 0.313 & 0.251 & 0.416  \\    
 \hline
\hline
\end{tabular}
\caption{Result comparison for different cross-validation method in BC1 dataset}
\label{table:stable3}
\end{table}

In the case of EGN, factors 2) and 3) — different normalization methods and variations in metric calculations — significantly contribute to the performance gap observed. Our approach to normalization for ST data involves dividing each gene's count by the total expression count of each spot and applying a log transformation, complemented by the expression smoothing method proposed by ST-Net \cite{he2020integrating}. Conversely, EGN's methodology adds a pseudo count of 1, applies a log transformation, and then conducts min-max normalization using each gene's max and min count values from all training data. We hypothesize that EGN's approach may inadvertently amplify technical variations due to batch effects, diminishing the focus on biologically relevant variations, which is central to our study. Therefore, we opt for a normalization method that we believe better preserves these biological variations.
Regarding evaluation methods, as detailed in Section 1, our approach involves predicting gene expression levels for each slide and averaging the outcomes across multiple slides. In contrast, EGN evaluates all spots of the validation data in a single assessment. Given the clinical context where a WSI is typically provided for gene expression prediction, we find our method more aligned with real-world applications. To further explore these methodological differences, we conduct experiments substituting our methods with those of EGN (s/ norm, s/ eval, and both combined as s/ norm\&eval). The experiment results, depicted in Table \ref{table:stable4}, confirm that the original results reported in EGN's literature are reproducible when adopting their specific normalization and evaluation strategies. 

\begin{table}[ht]
\begin{adjustbox}{width=0.47\textwidth}
\begin{tabular}{c|cccc}
\noalign{\smallskip}\noalign{\smallskip}\hline\hline
Model & \multicolumn{4}{c}{EGN} \\
    & MSE & MAE & PCC(M) & PCC(H) \\
\hline
Ours & 0.1923 & 0.3366 & 0.1112 & 0.2025 \\
s/ eval & 0.1930 & 0.3365 & 0.1494 & 0.3056  \\       
s/ norm & 0.0005 & 0.0173 & 0.1595 & 0.2193  \\    
s/ eval\&norm \cite{yang2023exemplar} & 0.0003 & 0.0134 & 0.2003 & 0.3011  \\    
 \hline
\hline
\end{tabular}
\end{adjustbox}
\caption{Result comparison for different evaluation and normalization method in BC2 dataset.}
\label{table:stable4}
\end{table}

In summary, our findings demonstrate that the discrepancies observed between our experimental results and those reported in existing literature arise primarily from differences in methodological approaches, particularly in cross-validation, normalization, and evaluation metrics, rather than from a lack of extensive hyperparameter tuning. These results highlight the critical impact of methodological choices in computational biology and the need for meticulous methodological reporting to ensure accurate comparisons and reproducibility. 

\subsection{Additional Ablation Studies}
We conduct further ablation studies on the BC1, BC2, and Visium datasets, with the results detailed in Tables \ref{table:stable5}, \ref{table:stable6}, and \ref{table:stable7} for each dataset respectively. In these studies, we examine the impact of different components of our model to understand their individual contributions to performance. A consistent trend observed across all datasets, aligning with findings from the SCC dataset, is the pronounced significance of the GEM. The GEM, central to our model's architecture, has shown to be particularly influential in enhancing performance.

\begin{table}[ht]
\begin{adjustbox}{width=0.45\textwidth}
\begin{tabular}{c|cccc}
\noalign{\smallskip}\noalign{\smallskip}\hline\hline
Dataset & \multicolumn{4}{c}{BC1} \\
    & MSE & MAE & PCC(M) & PCC(T) \\
\hline
 w/o TEM & 0.229 & 0.363 & 0.315 & 0.501 \\ 
 w/o NEM & 0.240 & 0.372 & 0.295 & 0.478 \\
 w/o GEM & 0.228 & 0.362 & 0.266 & 0.448 \\
\hline
 w/o PEG & 0.227 & 0.363 & 0.294 & 0.466 \\
 PEG & 0.230 & 0.365 & 0.304 & 0.485 \\
 \hline
  Summation & 0.241 & 0.375 & 0.293 & 0.475 \\
Concatenation & 0.239 & 0.372 & 0.297 & 0.484 \\
Attentional fusion & 0.237 & 0.370 & 0.311 & 0.502 \\
\hline
 w/o fusion loss & 0.246 & 0.377 & 0.295 & 0.481 \\    
 \rowcolor{beaublue}
 \hline
  Ours & 0.228 & 0.362 & 0.314 & 0.497 \\
\hline
\hline
\end{tabular}
\end{adjustbox}
\caption{Ablation studies in BC1 dataset}
\label{table:stable5}
\end{table}

\begin{table}[ht]
\begin{adjustbox}{width=0.45\textwidth}
\begin{tabular}{c|cccc}
\noalign{\smallskip}\noalign{\smallskip}\hline\hline
Dataset & \multicolumn{4}{c}{BC2} \\
    & MSE & MAE & PCC(M) & PCC(T) \\
\hline
 w/o TEM & 0.203 & 0.346 & 0.208 & 0.356 \\ 
 w/o NEM & 0.202 & 0.344 & 0.199 & 0.347 \\
 w/o GEM & 0.193 & 0.336 & 0.159 & 0.291 \\
\hline
 w/o PEG & 0.192 & 0.335 & 0.194 & 0.341 \\
 PEG & 0.196 & 0.338 & 0.201 & 0.350 \\
 \hline
  Summation & 0.203 & 0.346 & 0.186 & 0.335 \\
Concatenation & 0.205 & 0.346 & 0.190 & 0.337 \\
Attentional fusion & 0.198 & 0.340 & 0.198 & 0.354  \\
\hline
 w/o fusion loss & 0.211 & 0.349 & 0.203 & 0.355 \\    
 \rowcolor{beaublue}
 \hline
  Ours & 0.202 & 0.343 & 0.206 & 0.352 \\
\hline
\hline
\end{tabular}
\end{adjustbox}
\caption{Ablation studies in BC2 dataset}
\label{table:stable6}
\end{table}

\begin{table}[ht]
\begin{adjustbox}{width=0.45\textwidth}
\begin{tabular}{c|cccc}
\noalign{\smallskip}\noalign{\smallskip}\hline\hline
Dataset & \multicolumn{4}{c}{10X Visium} \\
    & MSE & MAE & PCC(M) & PCC(T) \\
\hline
 w/o TEM & 0.338 & 0.453 & 0.099 & 0.250 \\ 
 w/o NEM & 0.322 & 0.443 & 0.107 & 0.238 \\
 w/o GEM & 0.325 & 0.439 & 0.087 & 0.237 \\
\hline
 w/o PEG & 0.339 & 0.455 & 0.087 & 0.264 \\
 PEG & 0.371 & 0.475 & 0.109 & 0.248 \\
 \hline
  Summation & 0.342 & 0.451 & 0.106 & 0.225 \\
Concatenation & 0.332 & 0.446 & 0.089 & 0.232 \\
Attentional fusion & 0.327 & 0.447 & 0.110 & 0.267 \\
\hline
 w/o fusion loss & 0.328 & 0.442 & 0.050 & 0.206  \\    
 \rowcolor{beaublue}
 \hline
  Ours & 0.306 & 0.427 & 0.136 & 0.293 \\
\hline
\hline
\end{tabular}
\end{adjustbox}
\caption{Ablation studies in Visium dataset}
\label{table:stable7}
\end{table}

\subsection{Detailed Hyperparameter Settings}
In our study, hyperparameter tuning for each dataset is meticulously conducted using the WanDB platform \cite{wandb}. We set the range of hyperparameters based on the defaults reported in relevant literature, as shown in Table \ref{table:stable8}. 
For each model and dataset combination, we undertake a minimum of 100 experiments to determine the optimal settings.

The hyperparameters for each baseline model are detailed in their respective publications \cite{pang2021leveraging,zeng2022spatial,yang2023exemplar,xie2023spatially}. Regarding TRIPLEX, 'depth1', 'depth2', and 'depth3' refer to the depths of the transformer blocks in the Fusion Layer, Global Encoder, and Neighbor Encoder, respectively. 'num\_heads' denotes the number of heads in the multi-head self-attention mechanism of each transformer block, the 'mlp\_ratio' is the ratio of the MLP dimension to the embedding dimension within the transformer's FeedForward network, and 'dropout' represents the dropout probability in Transformer block. These hyperparameters are fine-tuned to maximize the PCC(M).
The final hyperparameters, as determined through our extensive experiments, are presented in Table \ref{table:stable9}.

 \begin{table}[ht]
    \begin{adjustbox}{width=0.48\textwidth}
    \begin{tabular}{|l|l|l|l|}
        \hline
        Model & \multicolumn{3}{c|}{HisToGene}  \\
        \hline
        \textbf{Parameter} & \textbf{Distribution} & \textbf{Min/Values} & \textbf{Max} \\
        \hline
        n\_layers & int\_uniform & 2 & 8 \\
        \hline
        dim & categorical & \multicolumn{2}{c|}{512,1024,2048} \\
        \hline
        num\_heads & categorical & \multicolumn{2}{c|}{4,8,16,32} \\
        \hline
        dropout & categorical & \multicolumn{2}{c|}{0.1,0.2,0.3,0.4} \\
        \hline
        \hline
        Model & \multicolumn{3}{c|}{HisT2ST}  \\
        \hline
        \textbf{Parameter} & \textbf{Distribution} & \textbf{Min/Values} & \textbf{Max} \\
        \hline
        depth1 & int\_uniform & 1 & 4 \\
        \hline
        depth2 & int\_uniform & 2 & 8 \\
        \hline
        depth3 & int\_uniform & 1 & 4 \\
        \hline
        heads & categorical & \multicolumn{2}{c|}{4,8,16} \\
        \hline
        channel & categorical & \multicolumn{2}{c|}{16,32,64,128} \\
        \hline
        bake & categorical & \multicolumn{2}{c|}{3,5,7} \\
        \hline
        kernel\_size & categorical & \multicolumn{2}{c|}{3,5,7} \\
        \hline
        \hline
        Model & \multicolumn{3}{c|}{EGN}  \\
        \hline
        \textbf{Parameter} & \textbf{Distribution} & \textbf{Min/Values} & \textbf{Max} \\
        \hline
        dim & categorical & \multicolumn{2}{c|}{512,1024,2048} \\
        \hline
        mlp\_dim & categorical & \multicolumn{2}{c|}{1024,2048,4096} \\
        \hline
        depth & categorical & \multicolumn{2}{c|}{2,4,6,8} \\
        \hline
        heads & categorical & \multicolumn{2}{c|}{4,8,16} \\
        \hline
        bhead & categorical & \multicolumn{2}{c|}{4,8,16} \\
        \hline
        bdim & categorical & \multicolumn{2}{c|}{32,64,128} \\
        \hline
        \hline
        Model & \multicolumn{3}{c|}{BLEEP}  \\
        \hline
        \textbf{Parameter} & \textbf{Distribution} & \multicolumn{2}{c|}{\textbf{Values}} \\
        \hline
        projection\_dim & categorical & \multicolumn{2}{c|}{128,256,512,1024,2048} \\
        dropout & categorical & \multicolumn{2}{c|}{0.1,0.15,0.2,0.25,0.3,0.35,0.4} \\
        \hline
        \hline
        Model & \multicolumn{3}{c|}{TRIPLEX}  \\
        \hline
        \textbf{Parameter} & \textbf{Distribution} & \textbf{Min/Values} & \textbf{Max} \\
        \hline
        depth1 & int\_uniform & 1 & 4 \\
        \hline
        depth2 & int\_uniform & 2 & 4 \\
        \hline
        depth3 & int\_uniform & 1 & 4 \\
        \hline
        dropout & categorical & \multicolumn{2}{c|}{0.1,0.2,0.3,0.4} \\
        \hline
        mlp\_ratio & categorical & \multicolumn{2}{c|}{1,2,4} \\
        \hline
        num\_heads & categorical & \multicolumn{2}{c|}{4,8,16} \\
        \hline
    \end{tabular}
    \end{adjustbox}
    \caption{Hyperparameters to be tuned}
    \label{table:stable8}
\end{table}

 \begin{table}[h]
    \centering
    \begin{tabular}{|l|l|l|l|}
        \hline
        Model & \multicolumn{3}{c|}{HisToGene}  \\
        \hline
        \textbf{Parameter} & \textbf{BC1} & \textbf{BC2} & \textbf{SCC} \\
        \hline
        n\_layers & 4 & 3 & 5 \\
        \hline
        dim & 2048 & 2048 & 512 \\
        \hline
        num\_heads & 4 & 16 & 4 \\
        \hline
        dropout & 0.4 & 0.3 & 0.4 \\
        \hline
        \hline
        Model & \multicolumn{3}{c|}{HisT2ST}  \\
        \hline
        \textbf{Parameter} & \textbf{BC1} & \textbf{BC2} & \textbf{SCC} \\
        \hline
        depth1 & 4 & 2 & 2 \\
        \hline
        depth2 & 3 & 6 & 6 \\
        \hline
        depth3 & 4 & 1 & 2 \\
        \hline
        heads & 8 & 4 & 8 \\
        \hline
        channel & 64 & 32 & 16 \\
        \hline
        bake & 3 & 5 & 7 \\
        \hline
        kernel\_size & 3 & 3 & 7 \\
        \hline
        \hline
        Model & \multicolumn{3}{c|}{EGN}  \\
        \hline
        \textbf{Parameter} & \textbf{BC1} & \textbf{BC2} & \textbf{SCC} \\
        \hline
        dim & 2048 & 512 & 2048 \\
        \hline
        mlp\_dim & 2048 & 4096 & 2048 \\
        \hline
        depth & 6 & 6 & 8 \\
        \hline
        heads & 4 & 8 & 16 \\
        \hline
        bhead & 16 & 4 & 16 \\
        \hline
        bdim & 128 & 64 & 64 \\
        \hline
        \hline
        Model & \multicolumn{3}{c|}{BLEEP}  \\
        \hline
        \textbf{Parameter} & \textbf{BC1} & \textbf{BC2} & \textbf{SCC} \\
        \hline
        projection\_dim & 128 & 128 & 128 \\
        \hline
        dropout & 0.4 & 0.3 & 0.35 \\
        \hline
        \hline
        Model & \multicolumn{3}{c|}{TRIPLEX}  \\
        \hline
        \textbf{Parameter} & \textbf{BC1} & \textbf{BC2} & \textbf{SCC} \\
        \hline
        depth1 & 1 & 3 & 2 \\
        \hline
        depth2 & 3 & 3 & 2 \\
        \hline
        depth3 & 3 & 4 & 4 \\
        \hline
        dropout1 & 0.2 & 0.4 & 0.1 \\
        \hline
        dropout2 & 0.1 & 0.1 & 0.1 \\
        \hline
        dropout3 & 0.3 & 0.3 & 0.3 \\
        \hline
        mlp\_ratio1 & 4 & 4 & 4 \\
        \hline
        mlp\_ratio2 & 4 & 2 & 1 \\
        \hline
        mlp\_ratio3 & 1 & 4 & 1 \\
        \hline
        num\_heads1 & 4 & 16 & 8 \\
        \hline
        num\_heads2 & 16 & 8 & 16 \\
        \hline
        num\_heads3 & 16 & 8 & 16 \\
        \hline
    \end{tabular}
    \caption{Selected hyperparameters in each dataset}
    \label{table:stable9}
\end{table}

\section{Additional Visualizations}
\begin{figure*}[!t]
    \centering
    \includegraphics[width=1\linewidth,height=9.5cm]{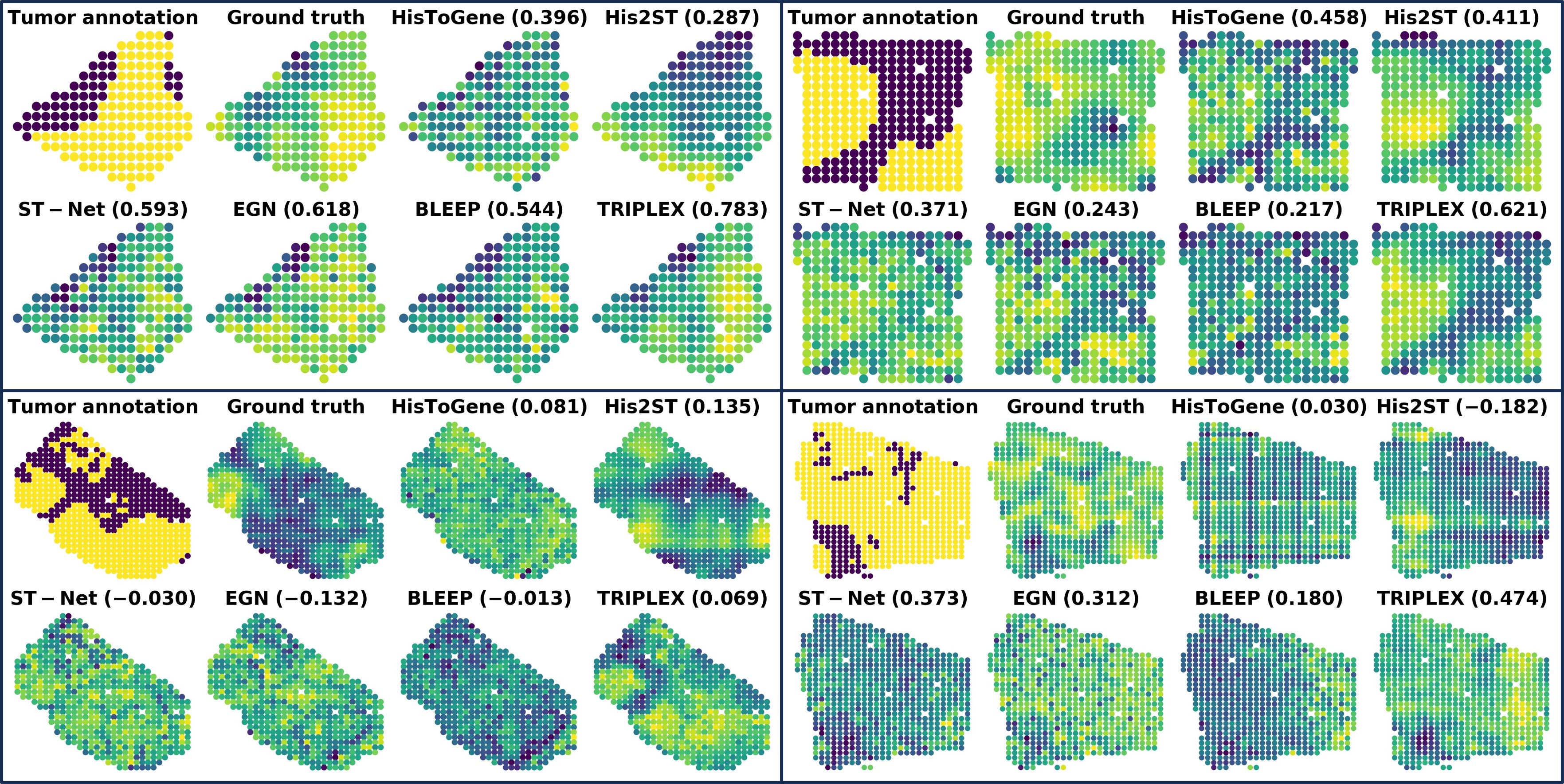}
    \caption{Additional visualization for predicting GNAS gene expression levels in BC1 dataset. We display the Pearson Correlation Coefficient (PCC) values between the ground truth and the prediction of the GNAS expression level estimated by each model.}
    \label{fig:sFigure8}
\end{figure*}

\begin{figure*}[!t]
    \centering
    \includegraphics[width=1\linewidth,height=9.5cm]{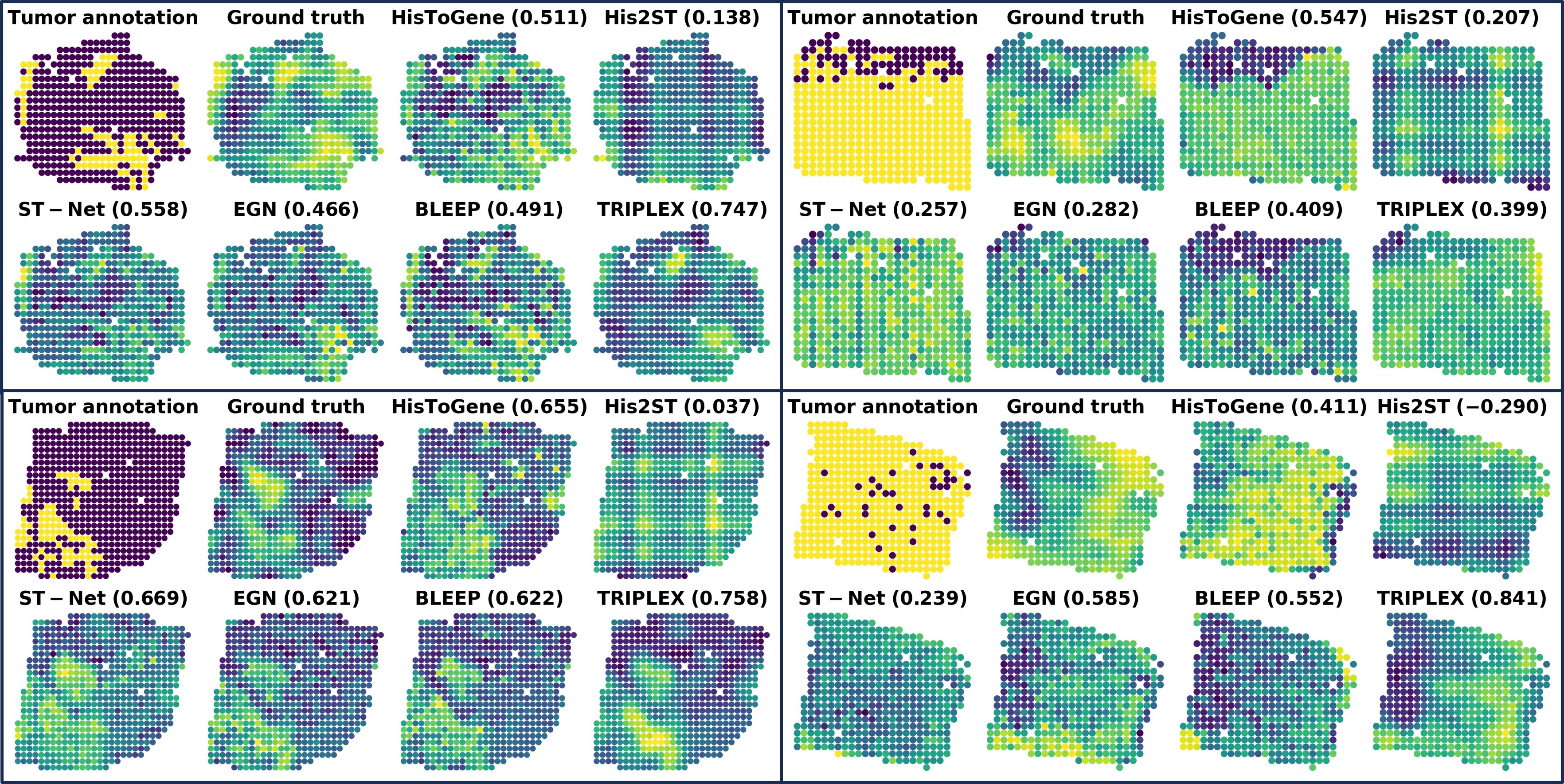}
    \caption{Additional visualization for predicting GNAS gene expression levels in BC2 dataset.}
    \label{fig:sFigure9}
\end{figure*}

\begin{figure*}[!t]
    \centering
    \includegraphics[width=1\linewidth]{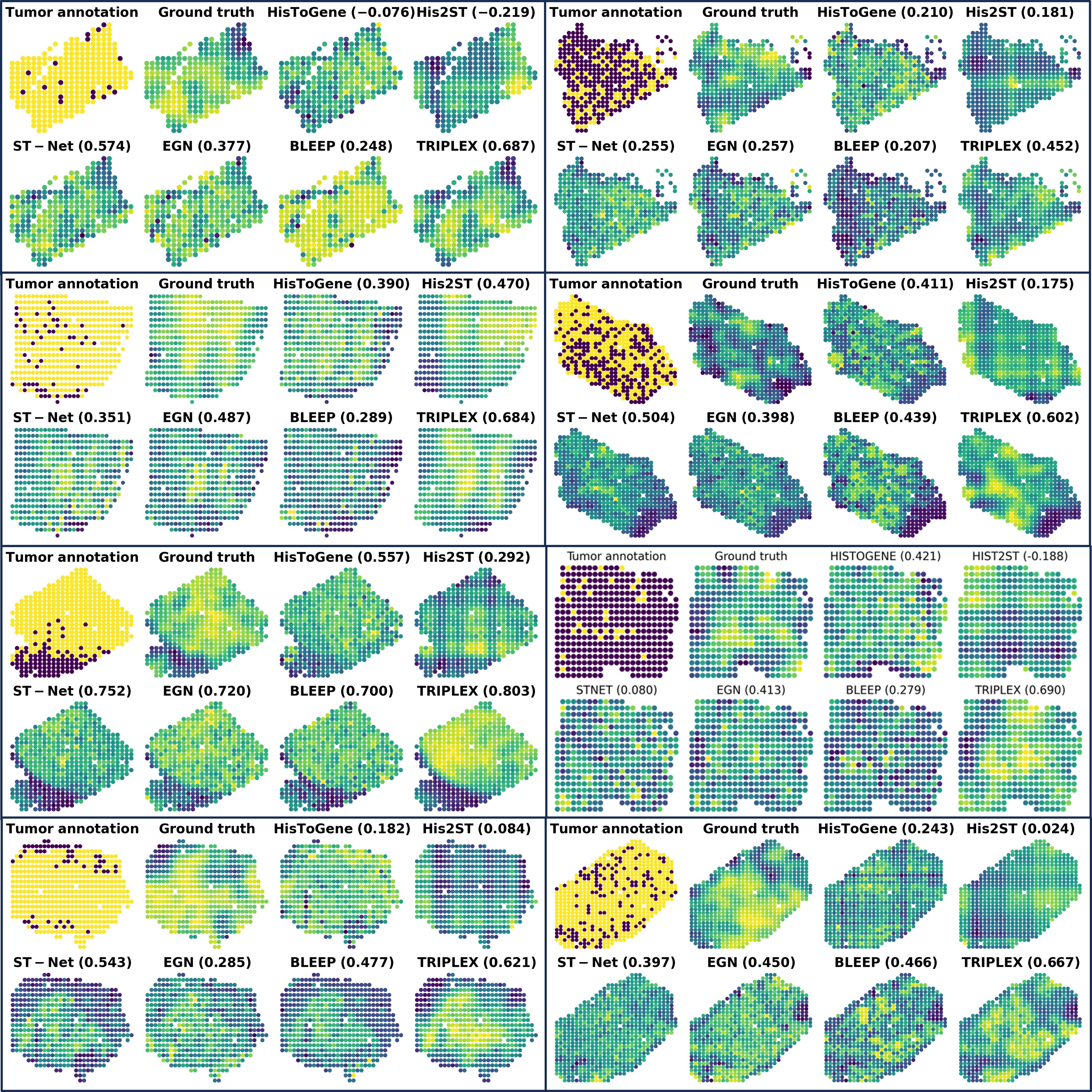}
    \caption{Additional visualization for predicting GNAS gene expression levels in BC2 dataset.}
    \label{fig:sFigure10}
\end{figure*}

\begin{figure*}[!t]
    \centering
    \includegraphics[width=1\linewidth]{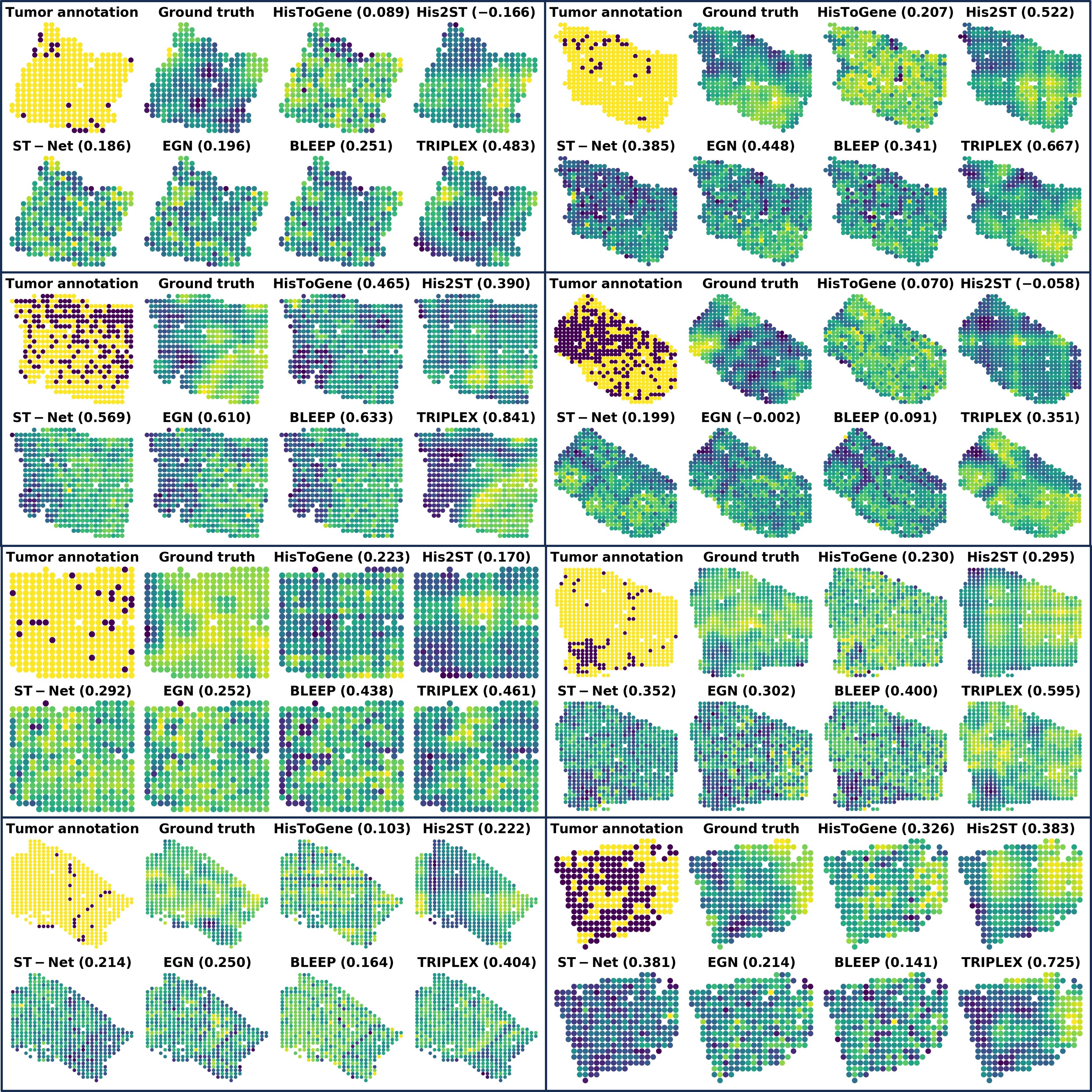}
    \caption{Additional visualization for predicting GNAS gene expression levels in BC2 dataset.}
    \label{fig:sFigure11}
\end{figure*}

In this section, we present additional visualizations focusing on the spatial expression distribution prediction of the GNAS gene, as shown in Figures \ref{fig:sFigure8} to \ref{fig:sFigure11}. These visualizations include four additional samples from the BC1 dataset and 20 samples from the BC2 dataset.
In analyzing these visualizations, we observe a high degree of consistency between the GNAS expression distribution and the annotations provided by pathologists. Notably, the predictions made by TRIPLEX demonstrate a markedly high accuracy, as quantitatively assessed against benchmark metrics, and align closely with tumor annotations. This consistency is particularly evident when compared to the predictions from other baseline models \cite{he2020integrating,pang2021leveraging,zeng2022spatial,yang2023exemplar,xie2023spatially}.

\end{document}